\documentclass[runningheads]{llncs}

\usepackage{eccv}

\newif\ifsupplementary
\supplementaryfalse

\newif\ifcomments
\commentsfalse 

\newcommand{\collab}[3]{%
  \ifcomments
    [{\textbf{\color{#2} #1: #3}}]%
  \fi
}

\newcommand{\nergis}[1]{\collab{Nergis}{green!50!black}{#1}}
\newcommand{\hesam}[1]{\collab{Hesam}{orange}{#1}}
\newcommand{\remove}[1]{\collab{CanRemove}{gray!80}{#1}}
\newcommand{\movesupp}[1]{\collab{MoveSupp}{brown!80}{#1}}

\newcommand{\smallpm}[1]{\mathbin{{\scriptstyle\pm\,#1}}}

\definecolor{BetterGreen}{RGB}{0,130,0}
\definecolor{BetterRed}{RGB}{200,0,0}
\definecolor{LightGreen}{RGB}{215,255,215}
\definecolor{LightRed}{RGB}{255,215,215}

\newcommand{\toymodel}{{order-invariant MLP}}
\newcommand{\Toymodel}{{Order-invariant MLP}}

\newcommand{\filtervalues}{{density value}}
\newcommand{\Filtervalues}{{Density value}}

\newcommand{\originalcase}{{baseline}}
\newcommand{\Originalcase}{{Baseline}}

\newcommand{\eigvalues}{{Harris eigenvalues}}
\newcommand{\Eigvalues}{{Harris Eigenvalues}}

\newcommand{\bothshortvalues}{{Eig.~+ Density}}

\newcommand{\bothfullvalues}{{Harris eigenvalues and density value}}

\newcommand{\cameraheight}{H}
\newcommand{\camerawidth}{W}
\newcommand{\densitybasedOMAC}{\mathcal{O}(w_d^2)}
\newcommand{\cornerbasedOMAC}{\mathcal{O}(w_d^2\,\texttt{ksize}^2)}

\newif\ifshowadded
\showaddedtrue
\ifshowadded

\else

\fi

\usepackage{eccvabbrv}

\usepackage{graphicx}
\usepackage{booktabs}
\usepackage{multirow}
\usepackage{adjustbox}
\usepackage{graphicx}
\usepackage{subcaption}
\usepackage{algorithm}
\usepackage{algpseudocode}
\usepackage{colortbl}
\usepackage{amsmath}
\usepackage{amssymb}
\usepackage{amsfonts}
\usepackage{xcolor}
\usepackage{wrapfig}
\usepackage{bibunits}

\usepackage[accsupp]{axessibility}  

\usepackage{hyperref}

\usepackage{orcidlink}

\begin{document}

\ifsupplementary
\title{Supplementary Material:\texorpdfstring{\\}{ - }{\small Static in Frames, Dynamic in Events: Rethinking Features in Event Cameras as Motion Cues}}
\else
\title{Static in Frames, Dynamic in Events: Rethinking Features in Event Cameras as Motion Cues}
\fi

\ifsupplementary
\titlerunning{Supplementary Material}
\else
\titlerunning{Event Features for Motion Cues}
\fi

\author{Hesam Araghi\orcidlink{0000-0002-4539-4408} \and
Jan van Gemert\orcidlink{0000-0002-3913-2786} \and
Nergis Tomen\orcidlink{0000-0003-3916-1859}}

\authorrunning{H.~Araghi et al.}

\institute{
Computer Vision Lab, Delft University of Technology\\
\email{\{h.araghi, j.c.vangemert, n.tomen\}@tudelft.nl}}

\maketitle

\makeatletter
\begin{bibunit}[splncs04]
\ifsupplementary
\appendix

\section{Computing $A_1$ and $A_2$ from Eigenvalues}
\label{subsec:compute_A1_A2}

This subsection provides the derivation of how to recover $A_1$ and $A_2$
from the eigenvalues of the structure tensor.
Starting from the structure-tensor formulation in the main paper, we can write the structure tensor as

\begin{equation}
    S_w = V
\begin{bmatrix}
    A_1 & 0 \\
    0 & A_2
\end{bmatrix}
V^T.
\end{equation}
The trace of $S_w$ is given by
\begin{align}
\mathrm{trace}(S_w)
&= \mathrm{trace}\!\left(V
\begin{bmatrix}
A_1 & 0 \\
0 & A_2
\end{bmatrix}
V^T\right) \nonumber\\
&= \mathrm{trace}\!\left(
\begin{bmatrix}
A_1 & 0 \\
0 & A_2
\end{bmatrix}
V^T V
\right) \nonumber\\
&= \mathrm{trace}\!\left(
\begin{bmatrix}
A_1 & 0 \\
0 & A_2
\end{bmatrix}
\begin{bmatrix}
1 & -\cos(\theta) \\
-\cos(\theta) & 1
\end{bmatrix}
\right) \nonumber\\
&= A_1 + A_2.
\label{eq:trace}
\end{align}

Similarly, the determinant of $S_w$ is
\begin{align}
\det(S_w)
&= \det\!\left(V
\begin{bmatrix}
A_1 & 0 \\
0 & A_2
\end{bmatrix}
V^T\right) \nonumber\\
&= \det\!\left(
\begin{bmatrix}
A_1 & 0 \\
0 & A_2
\end{bmatrix}
\begin{bmatrix}
1 & -\cos(\theta) \\
-\cos(\theta) & 1
\end{bmatrix}
\right) \nonumber\\
&= A_1 A_2 \big(1 - \cos^2(\theta)\big) \nonumber\\
&= A_1 A_2 \sin^2(\theta).
\label{eq:det}
\end{align}

From Eqs.~\eqref{eq:trace} and \eqref{eq:det}, we observe that
\begin{align}
    A_1 + A_2 &= \mathrm{trace}(S_w) = \lambda_1 + \lambda_2,\label{eq:sum of As}\\
A_1 A_2 &= \frac{\det(S_w)}{\sin^2(\theta)} = \frac{\lambda_1 \lambda_2}{\sin^2(\theta)}.\label{eq:prod of As}
\end{align}
Therefore, $A_1$ and $A_2$ are the two roots of the quadratic equation
\begin{equation}
A^2 - (\lambda_1 + \lambda_2)A + \frac{\lambda_1 \lambda_2}{\sin^2(\theta)} = 0.
\label{eq:quadratic supp}
\end{equation}

\section{IDNet}
\label{supp:subsec:idnet}

\subsection{IDNet Architecture Details}
\label{supp:subsubsec:idnet_architecture}

IDNet uses a voxel grid representation of events as input.
Given a time window, events are binned into $B$ temporal bins to form a tensor of shape $B \times H \times W$, where $H$ and $W$ are the spatial dimensions.
A convolutional encoder extracts features at a reduced spatial resolution ($1/D_s$ resolution).
The encoder output, with dimension \texttt{rnn\_input\_dim}, is fed into a recurrent update block based on a ConvGRU with hidden dimension \texttt{hidden\_dim}.
The network iteratively refines the flow estimate over $N_{\text{iter}}$ deblurring iterations.
At each iteration, the current flow estimate is used to warp the event representation, and the update block predicts a residual flow correction.
The final flow is upsampled to full resolution using a learned convex upsampling mask.

\subsection{Tiny IDNet}
\label{supp:subsubsec:tiny_idnet}

For the toy dataset experiments, we use a reduced-complexity variant of IDNet, which we refer to as \emph{Tiny IDNet}.
The overall iterative deblur-and-refine architecture remains identical to the full IDNet described in \cref{supp:subsubsec:idnet_architecture}; the changes target only the channel widths and encoder depth to obtain a lightweight model suitable for the simplicity of the toy setting.
The following modifications are applied:

\begin{enumerate}
    \item \textbf{Encoder depth.}
    Each residual layer in the feature encoder uses a single residual block instead of two, effectively halving the encoder depth (2 residual blocks total instead of 4).

    \item \textbf{RNN input dimension.}
    The RNN input dimension (\texttt{rnn\_input\_dim}) is reduced, which proportionally shrinks the encoder output.

    \item \textbf{Hidden dimension.}
    The ConvGRU hidden dimension (\texttt{hidden\_dim}) is reduced, which also reduces the flow head dimensions accordingly.

    \item \textbf{Upsampling mask head.}
    The intermediate channel count in the learned convex upsampling mask network is reduced from 256 to 32.
\end{enumerate}

\noindent All other design choices---the ConvGRU recurrence, the convex upsampling mechanism, the deblur-and-iterate loop, and the 1/8 resolution ($D_s{=}8$)---remain unchanged.
We use $B{=}5$ temporal bins and $N_{\text{iter}}{=}4$ deblurring iterations for the toy dataset experiments.

\subsection{IDNet Training Details}
\label{supp:subsubsec:idnet_training}

To incorporate our proposed features, we voxelize the per-event \filtervalues{} $I$ and \eigvalues{} $(\lambda_1, \lambda_2)$ into the same temporal bins as the event voxel grid.
These feature volumes are concatenated along the feature dimension, resulting in an input tensor of shape $B \times F \times H \times W$, where $F \in \{1, 2, 3, 4\}$ corresponds to baseline events, \filtervalues{}, \eigvalues{}, and \bothshortvalues{} configurations, respectively.
The first convolutional layer of the encoder is modified to match the extended input feature dimension.
We use $B=15$ temporal bins for DSEC and $B=5$ for toy datasets, at 1/4 and 1/8 resolutions ($D_s \in \{4, 8\}$), $N_{\text{iter}}=4$ deblurring iterations, and train using the Adam optimizer (learning rate $10^{-4}$), a OneCycleLR scheduler, and sparse $\ell_1$ loss.

To compute the \filtervalues{} features for IDNet and Tiny IDNet, we use a spatial filter size of $5 \times 5$ and a temporal decay of $\tau=1$~ms.

\subsection{Ablation on Filter Size and Temporal Decay}
\label{supp:subsec:filter_tau_ablation}

\Cref{tab:ablation} reports the endpoint error (EPE) of IDNet for different density-filter sizes and temporal decay values $\tau$, trained on a subset of DSEC and evaluated on the validation sequence \texttt{zurich\_city\_01\_a}.
The results show robustness across a range of $\tau$ values and filter sizes.
Based on this ablation, we use a filter size of 5 and $\tau = 1$~ms for DSEC.

\begin{table}[h]
\centering
\caption{Ablation study on the density-filter size and temporal decay $\tau$ using IDNet, reporting EPE on the DSEC validation sequence \texttt{zurich\_city\_01\_a}.}
\label{tab:ablation}
\begin{tabular}{c@{\hspace{7mm}}c@{\hspace{7mm}}c@{\hspace{7mm}}c}
\toprule
Filter Size & $\tau = 1\,ms$ & $\tau = 5\,ms$ & $\tau = 15\,ms$ \\
\midrule
5 & 0.786 & 0.804 & 0.793 \\
7 & 0.795 & 0.789 & 0.805 \\
\bottomrule
\end{tabular}
\end{table}

\section{Toy Dataset}
\label{supp:sec:toy_dataset}

\subsection{Additional Metrics for Different $K$ and Hidden Dimensions}
\label{supp:subsec:toy_additional_metrics}

\Cref{tab:toy:k5,tab:toy:k10,tab:toy:k50} report additional performance metrics (endpoint error, angular error, and percentage of outliers) for the \toymodel{} across different neighborhood sizes $K \in \{5, 10, 50\}$ and hidden dimensions 64 and 128, complementing the MSE loss curves shown in the main paper.

To compute the \filtervalues{} features for the toy dataset, we use a spatial filter size of $7 \times 7$ and a temporal decay of $\tau=30$~ms.

\subsection{Toy Dataset Visualizations}
\label{supp:sec:toy_dataset_visualization}

\subsection{Textured Conditions: Additional Metrics and Visualizations}
\label{supp:subsec:toy_texture}

\Cref{tab:toy:texture_comparison} provides additional metrics for the texture experiments described in the main paper.
\Cref{fig:supp:texture_visualization_1,fig:supp:texture_visualization_2} visualize the toy dataset under different foreground textures from DTD~\cite{cimpoi14describing}, showing the original texture, the rendered intensity frame, and the generated event polarity scatter.
\Cref{fig:toy:texture} shows the MSE loss curves over training epochs for both \originalcase{} and \bothshortvalues{} feature settings across different textures for the \toymodel{}, while \Cref{fig:toy:tinyidnet:texture} shows the corresponding textured experiment for \emph{Tiny IDNet}.

\subsection{Shot Noise: Additional Metrics and Visualizations}
\label{supp:subsec:toy_shot_noise}

Shot noise events were generated using the V2E simulator~\cite{hu2021v2e}.
\Cref{fig:supp:shot_noise_visualization} visualizes the event polarity scatter under different shot noise frequencies.
\Cref{tab:toy:shot_noise_comparison} provides additional metrics for the shot noise experiments described in the main paper.
\Cref{fig:toy:shot noise} shows the MSE loss curves comparing \originalcase{} and \bothshortvalues{} features across noise levels.

\section{DSEC Sub-dataset Sequences}
\label{subsec:dsec_subdataset_sequences}

For the sub-dataset experiments on DSEC, we use approximately one-sixth of the full training data.
The following four sequences are used for training:
\begin{itemize}
    \item \texttt{zurich\_city\_02\_d}
    \item \texttt{zurich\_city\_05\_b}
    \item \texttt{zurich\_city\_08\_a}
    \item \texttt{zurich\_city\_11\_a}
\end{itemize}
These sequences were selected to provide a diverse subset of the DSEC dataset while keeping the training data limited to evaluate the benefit of feature extension in data-scarce scenarios.

\section{Additional Evaluation on EVIMO2v2}
\label{supp:sec:evimo}

To further assess the proposed features on real-world data, we additionally evaluate IDNet~\cite{wu_lightweight_2024} on the EVIMO2v2 dataset~\cite{burner_evimo2_2022}.
EVIMO2v2 has dense flow ground-truth for independently moving objects and not for ego-motion alone. Second, the sequences are recorded indoors at short range, rather than by the large, smoothly varying ego-motion fields typical of driving data.
Ground-truth flow frames are generated every 50\,ms, and all evaluation is carried out at the full $640\times480$ sensor resolution without cropping.
We consider two training cases of the subset and full dataset.
The reduced subset uses 4 of the 20 training sequences, corresponding to roughly one sixth of the  training samples, while the full regime uses all 20; in both cases one sequence is held out for validation.
As in the DSEC experiments, we compare the \originalcase{} against extended input with the \bothshortvalues{} features.
For the reduced-data setting, we report the mean and standard deviation over three independent runs.

The results are reported in \cref{tab:evimo-combined}.
In the reduced regime the
extended features improve the metrics in optical flow estimation. 
 When training on the full dataset, however, the performance of the two representations becomes comparable, and the advantage of the extended features diminishes. 
This behavior is consistent with our discussion in the main paper and mirrors the trend observed on DSEC: with sufficient training data, the network can increasingly learn the  geometric information provided explicitly by the proposed features directly from the data, reducing the benefit of encoding this information in the input representation.
 These results further support the benefit of the proposed features when the amount of training data is limited.

\begin{table}[h]
\centering
\caption{Optical flow evaluation on the EVIMO2v2
dataset~\cite{burner_evimo2_2022} using IDNet, under a sub-dataset (4 of 20 sequences) and the full training set. 
Sub-dataset results are the mean $\pm$ standard deviation over 3 independently trained runs. Best results in \textbf{bold}.}
\label{tab:evimo-combined}
\begin{tabular}{l l cccc}
\hline
Category & Feature Type & $\mathrm{1PE}\downarrow$ & $\mathrm{3PE}\downarrow$ & $\mathrm{EPE}\downarrow$ & $\mathrm{AE}\downarrow$ \\
\hline
\multirow{2}{*}{Sub-dataset}
    & \Originalcase{}
        & $17.533\,\smallpm{0.149}$ & $4.596\,\smallpm{0.064}$ & $0.938\,\smallpm{0.006}$ & $20.007\,\smallpm{0.234}$ \\
    & w/ Eig. + Density
        & $\mathbf{17.302}\,\smallpm{0.224}$ & $\mathbf{4.432}\,\smallpm{0.052}$ & $\mathbf{0.910}\,\smallpm{0.006}$ & $\mathbf{19.478}\,\smallpm{0.162}$ \\
\hline
\multirow{2}{*}{Full dataset}
    & \Originalcase{}
        & $\mathbf{11.663}$ & $\mathbf{2.232}$ & $\mathbf{0.633}$ & $\mathbf{15.473}$ \\
    & w/ Eig. + Density
        & 11.860 & 2.264 & 0.638 & 15.701 \\
\hline
\end{tabular}

\end{table}

\section{Parameter and Computational Overhead}
\label{supp:sec:overhead}

As \cref{tab:num param time} shows, the number of parameters added to IDNet is
small, since only the number of input channels of the first convolutional layer
of the encoder is increased; consequently, the added inference time is also
small. Averaged over the 1773 frames of the EVIMO2v2 evaluation split at full
$640\times480$ resolution. Latency is measured around the forward pass only, at batch size~1 in \texttt{fp32}
after 25 discarded warm-up iterations, on a system with a single NVIDIA GeForce RTX 5090 GPU and 8-core AMD Ryzen 7 CPU.

For computing the features themselves,  \cref{tab:computational complexity} reports the memory usage and the number of multiply-accumulate operations (MACs) per event, where $w_d$ denotes the size of the density filter introduced in the main paper and \texttt{ksize} is the size of the Sobel filters.
As these features are designed to be causal and hardware-friendly, embedding their computation in the event camera hardware is a suitable direction for further speed improvement.

\begin{table}[h]
    \centering
    \caption{Number of parameters and inference time for IDNet with and
    without the extended features. Inference time is the mean $\pm$ standard
    deviation over the 1773 frames of the EVIMO2v2 evaluation split at full
    $640\times480$ resolution.}
    \label{tab:num param time}
\begin{tabular}{l@{\hspace{7mm}}c@{\hspace{7mm}}c}
\toprule
Feature type   & \# parameters & inference (ms/frame)\\
\midrule
   \Originalcase{} & 1.427\,M & $30.43 \pm 0.56$  \\
   w/ \bothshortvalues{} & 1.430\,M & $32.58 \pm 0.39$\\
\bottomrule
\end{tabular}
\end{table}

\begin{table}[h]
    \centering
    \caption{Memory usage and computational complexity (MACs/event) for computing the \filtervalues{} and \eigvalues{} features.}
    \label{tab:computational complexity}
\begin{tabular}{l@{\hspace{7mm}}c@{\hspace{7mm}}c}
\toprule
Feature type   & Memory & Computational complexity (MACs/event) \\
\midrule
w/ \Filtervalues{}  & $\mathcal{O}(\cameraheight\camerawidth)$  & $\densitybasedOMAC$\\
w/ \bothshortvalues{} & $\mathcal{O}(\cameraheight\camerawidth)$ & $\cornerbasedOMAC$ \\
\bottomrule
\end{tabular}
\end{table}

\begin{table*}[h]
\centering
\caption{Performance metrics for $K=5$ using the \toymodel{}. Comparing \Originalcase{}, \Filtervalues{}, \Eigvalues{}, and \bothshortvalues{} across hidden dimensions 64 and 128. Best values in \textbf{bold}.}
\label{tab:toy:k5}
\begin{adjustbox}{width=\textwidth}%

\begin{tabular}{@{}lcccccccc@{}}
\toprule
\multirow{2}{*}{Feature Type}       & \multicolumn{4}{c}{hidden dim = 64} & \multicolumn{4}{c}{hidden dim = 128} \\ \cmidrule(lr){2-5}\cmidrule(l){6-9} 
                                    & L1 {\small $(\times 10^{-1})$} & MSE {\small $(\times 10^{-2})$} & EPE {\small $(\times 10^{-1})$} & AE & L1 {\small $(\times 10^{-1})$} & MSE {\small $(\times 10^{-2})$} & EPE {\small $(\times 10^{-1})$} & AE \\ \midrule
\Originalcase{} & $2.600 \smallpm{0.003}$ & $9.010 \smallpm{0.010}$ & $4.004 \smallpm{0.004}$ & $110.1 \smallpm{0.3}$ & $2.574 \smallpm{0.001}$ & $8.920 \smallpm{0.010}$ & $3.963 \smallpm{0.002}$ & $108.8 \smallpm{0.2}$ \\
\Filtervalues{} & $1.890 \smallpm{0.022}$ & $5.600 \smallpm{0.100}$ & $2.963 \smallpm{0.030}$ & $64.2 \smallpm{1.2}$ & $1.701 \smallpm{0.012}$ & $4.820 \smallpm{0.060}$ & $2.670 \smallpm{0.018}$ & $55.6 \smallpm{0.6}$ \\
\Eigvalues{} & $1.642 \smallpm{0.011}$ & $4.250 \smallpm{0.040}$ & $2.566 \smallpm{0.015}$ & $47.6 \smallpm{0.6}$ & $1.459 \smallpm{0.009}$ & $3.500 \smallpm{0.040}$ & $2.286 \smallpm{0.014}$ & $40.8 \smallpm{0.2}$ \\
\bothshortvalues{} & $\mathbf{1.324} \smallpm{0.018}$ & $\mathbf{2.980} \smallpm{0.060}$ & $\mathbf{2.076} \smallpm{0.026}$ & $\mathbf{39.1} \smallpm{0.7}$ & $\mathbf{1.136} \smallpm{0.010}$ & $\mathbf{2.310} \smallpm{0.030}$ & $\mathbf{1.784} \smallpm{0.014}$ & $\mathbf{32.0} \smallpm{0.3}$ \\
\bottomrule
\end{tabular}
\end{adjustbox}
\end{table*}

\begin{table*}[h]
\centering
\caption{Performance metrics for $K=10$ using the \toymodel{}. Comparing \Originalcase{}, \Filtervalues{}, \Eigvalues{}, and \bothshortvalues{} across hidden dimensions 64 and 128. Best values in \textbf{bold}.}
\label{tab:toy:k10}
\begin{adjustbox}{width=\textwidth}%

\begin{tabular}{@{}lcccccccc@{}}
\toprule
\multirow{2}{*}{Feature Type}       & \multicolumn{4}{c}{hidden dim = 64} & \multicolumn{4}{c}{hidden dim = 128} \\ \cmidrule(lr){2-5}\cmidrule(l){6-9} 
                                    & L1 {\small $(\times 10^{-1})$} & MSE {\small $(\times 10^{-2})$} & EPE {\small $(\times 10^{-1})$} & AE & L1 {\small $(\times 10^{-1})$} & MSE {\small $(\times 10^{-2})$} & EPE {\small $(\times 10^{-1})$} & AE \\ \midrule
\Originalcase{} & $2.322 \smallpm{0.009}$ & $7.740 \smallpm{0.040}$ & $3.605 \smallpm{0.013}$ & $83.9 \smallpm{0.5}$ & $2.262 \smallpm{0.006}$ & $7.470 \smallpm{0.020}$ & $3.509 \smallpm{0.009}$ & $79.7 \smallpm{0.3}$ \\
\Filtervalues{} & $1.588 \smallpm{0.011}$ & $4.110 \smallpm{0.040}$ & $2.499 \smallpm{0.016}$ & $50.6 \smallpm{0.7}$ & $1.343 \smallpm{0.013}$ & $3.110 \smallpm{0.050}$ & $2.105 \smallpm{0.021}$ & $40.1 \smallpm{0.6}$ \\
\Eigvalues{} & $1.329 \smallpm{0.010}$ & $2.920 \smallpm{0.040}$ & $2.088 \smallpm{0.015}$ & $35.5 \smallpm{0.3}$ & $1.166 \smallpm{0.006}$ & $2.320 \smallpm{0.020}$ & $1.833 \smallpm{0.010}$ & $30.3 \smallpm{0.3}$ \\
\bothshortvalues{} & $\mathbf{1.092} \smallpm{0.027}$ & $\mathbf{2.060} \smallpm{0.080}$ & $\mathbf{1.716} \smallpm{0.042}$ & $\mathbf{29.2} \smallpm{0.8}$ & $\mathbf{0.905} \smallpm{0.012}$ & $\mathbf{1.480} \smallpm{0.030}$ & $\mathbf{1.418} \smallpm{0.019}$ & $\mathbf{23.1} \smallpm{0.4}$ \\
\bottomrule
\end{tabular}
\end{adjustbox}
\end{table*}

\begin{table*}[h]
\centering
\caption{Performance metrics for $K=50$ using the \toymodel{}. Comparing \Originalcase{}, \Filtervalues{}, \Eigvalues{}, and \bothshortvalues{} across hidden dimensions 64 and 128. Best values in \textbf{bold}.}
\label{tab:toy:k50}
\begin{adjustbox}{width=\textwidth}%

\begin{tabular}{@{}lcccccccc@{}}
\toprule
\multirow{2}{*}{Feature Type}       & \multicolumn{4}{c}{hidden dim = 64} & \multicolumn{4}{c}{hidden dim = 128} \\ \cmidrule(lr){2-5}\cmidrule(l){6-9} 
                                    & L1 {\small $(\times 10^{-1})$} & MSE {\small $(\times 10^{-2})$} & EPE {\small $(\times 10^{-1})$} & AE & L1 {\small $(\times 10^{-1})$} & MSE {\small $(\times 10^{-2})$} & EPE {\small $(\times 10^{-1})$} & AE \\ \midrule
\Originalcase{} & $1.374 \smallpm{0.013}$ & $3.230 \smallpm{0.050}$ & $2.159 \smallpm{0.019}$ & $38.6 \smallpm{0.8}$ & $1.258 \smallpm{0.016}$ & $2.800 \smallpm{0.060}$ & $1.978 \smallpm{0.024}$ & $34.0 \smallpm{0.6}$ \\
\Filtervalues{} & $0.936 \smallpm{0.020}$ & $1.620 \smallpm{0.070}$ & $1.473 \smallpm{0.031}$ & $24.6 \smallpm{0.9}$ & $0.737 \smallpm{0.015}$ & $1.040 \smallpm{0.030}$ & $1.155 \smallpm{0.021}$ & $18.0 \smallpm{0.6}$ \\
\Eigvalues{} & $0.881 \smallpm{0.015}$ & $1.460 \smallpm{0.040}$ & $1.389 \smallpm{0.025}$ & $23.2 \smallpm{0.4}$ & $0.721 \smallpm{0.005}$ & $1.020 \smallpm{0.020}$ & $1.134 \smallpm{0.009}$ & $18.5 \smallpm{0.3}$ \\
\bothshortvalues{} & $\mathbf{0.700} \smallpm{0.024}$ & $\mathbf{0.920} \smallpm{0.050}$ & $\mathbf{1.098} \smallpm{0.038}$ & $\mathbf{17.7} \smallpm{0.6}$ & $\mathbf{0.560} \smallpm{0.007}$ & $\mathbf{0.620} \smallpm{0.020}$ & $\mathbf{0.872} \smallpm{0.011}$ & $\mathbf{13.9} \smallpm{0.4}$ \\
\bottomrule
\end{tabular}
\end{adjustbox}
\end{table*}

\begin{table*}[h]
\centering
\caption{Performance under textured conditions. Comparing \Originalcase{} and \bothshortvalues{} features on two DTD texture patterns using the \toymodel{} (hidden dim 128, $K=50$). Best values in \textbf{bold}.}
\label{tab:toy:texture_comparison}
\begin{adjustbox}{width=\textwidth}%

\begin{tabular}{@{}lcccccccc@{}}
\toprule
\multirow{2}{*}{Feature Type}       & \multicolumn{4}{c}{Flecked 0103} & \multicolumn{4}{c}{Spiralled 0137} \\ \cmidrule(lr){2-5}\cmidrule(l){6-9} 
                                    & L1 {\small $(\times 10^{-1})$} & MSE {\small $(\times 10^{-2})$} & EPE {\small $(\times 10^{-1})$} & AE & L1 {\small $(\times 10^{-1})$} & MSE {\small $(\times 10^{-2})$} & EPE {\small $(\times 10^{-1})$} & AE \\ \midrule
\Originalcase{} & $2.408 \smallpm{0.006}$ & $7.990 \smallpm{0.030}$ & $3.796 \smallpm{0.009}$ & $95.8 \smallpm{0.4}$ & $2.150 \smallpm{0.009}$ & $6.440 \smallpm{0.050}$ & $3.358 \smallpm{0.013}$ & $87.6 \smallpm{0.7}$ \\
\bothshortvalues{} & $\mathbf{2.096} \smallpm{0.016}$ & $\mathbf{6.680} \smallpm{0.100}$ & $\mathbf{3.314} \smallpm{0.027}$ & $\mathbf{75.2} \smallpm{0.9}$ & $\mathbf{1.521} \smallpm{0.020}$ & $\mathbf{4.110} \smallpm{0.080}$ & $\mathbf{2.423} \smallpm{0.029}$ & $\mathbf{54.5} \smallpm{0.9}$ \\
\bottomrule
\end{tabular}
\end{adjustbox}
\end{table*}

\newcommand{\texW}{0.30\textwidth}
\begin{figure*}[h]
  \centering
  \begin{tabular}{@{}c@{\hskip 4pt}c@{\hskip 4pt}c@{}}
    \textbf{Original Texture} & 
    \parbox[b][0.1cm][c]{\texW}{\centering\textbf{Intensity Frame \\at $t{=}203$~ms}} & 
    \parbox[b][0.1cm][c]{\texW}{\centering\textbf{Events \\at $t{=}200$--$205$~ms}} \\[4pt]
    \parbox[b][\texW][c]{\texW}{\centering\includegraphics[width=\texW]{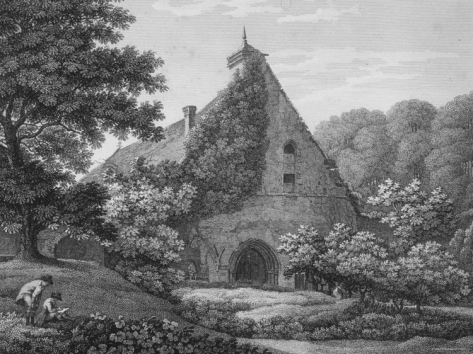}\\[2pt]{\small crosshatched}} &
    \includegraphics[width=\texW]{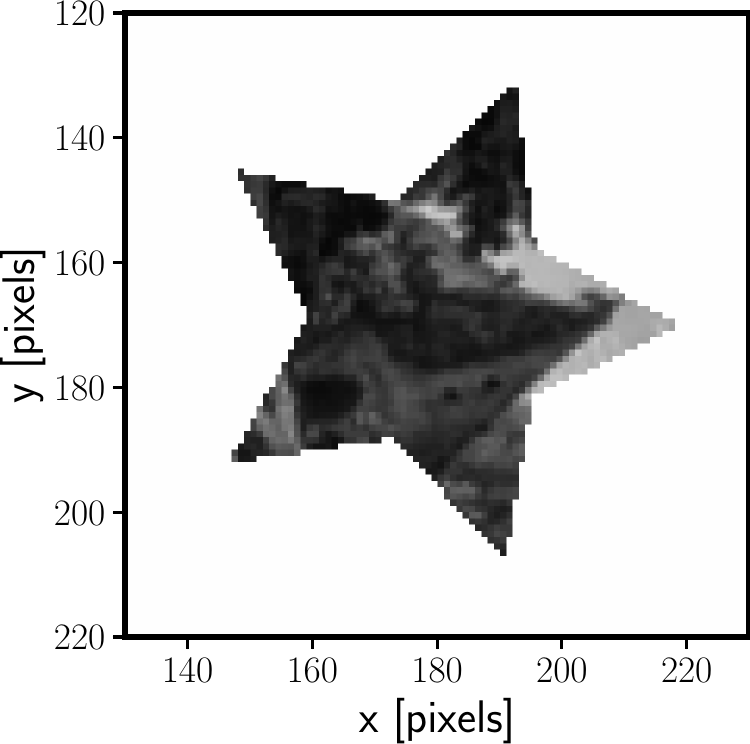} &
    \includegraphics[width=\texW]{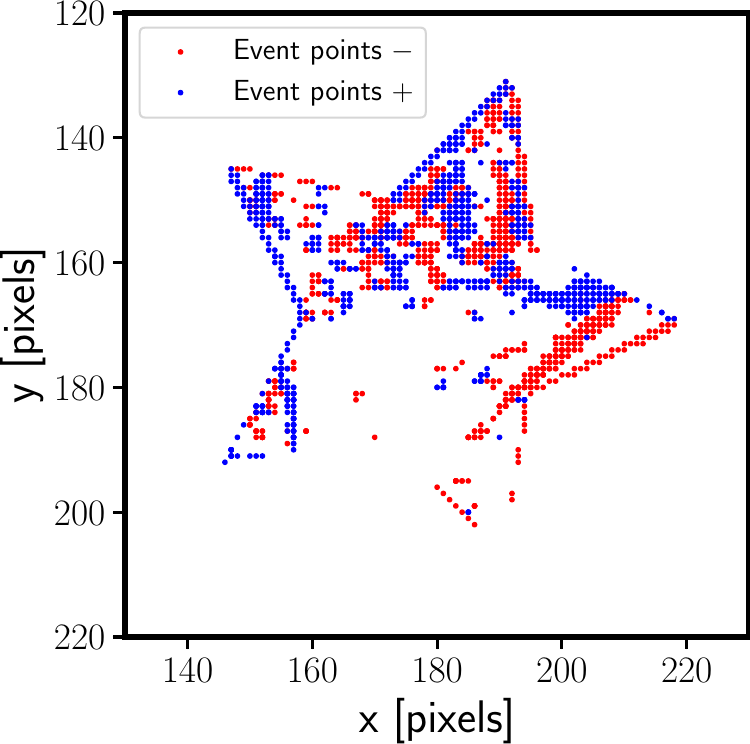} \\
    \parbox[b][\texW][c]{\texW}{\centering\includegraphics[height=0.25\textwidth]{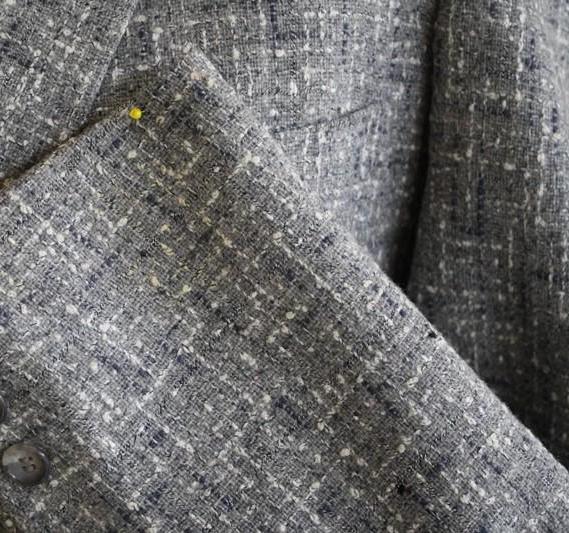}\\[2pt]{\small flecked}} &
    \includegraphics[width=\texW]{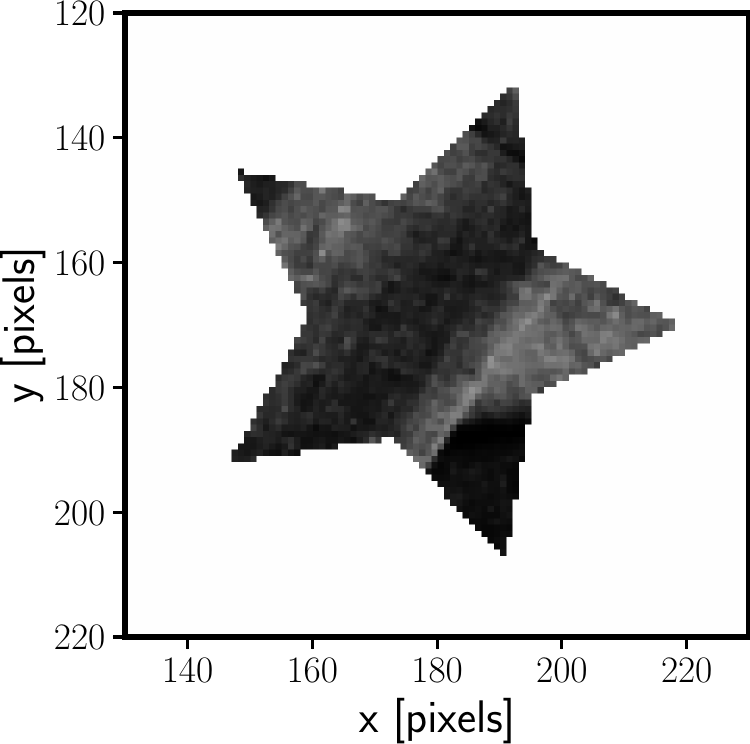} &
    \includegraphics[width=\texW]{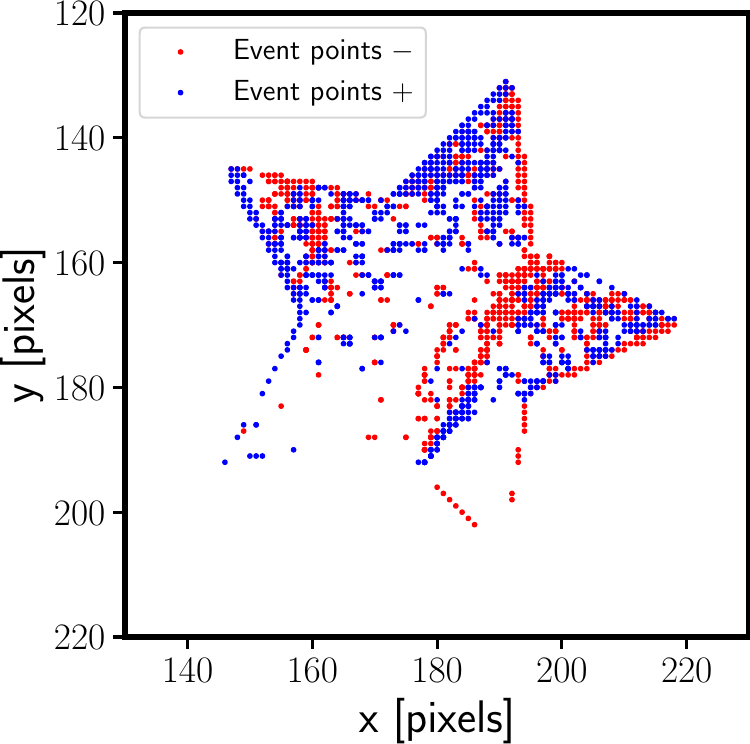} \\[4pt]
    \parbox[b][\texW][c]{\texW}{\centering\includegraphics[height=0.25\textwidth]{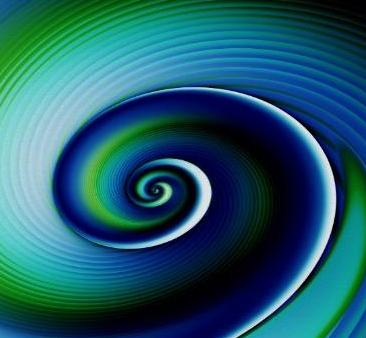}\\[2pt]{\small spiralled}} &
    \includegraphics[width=\texW]{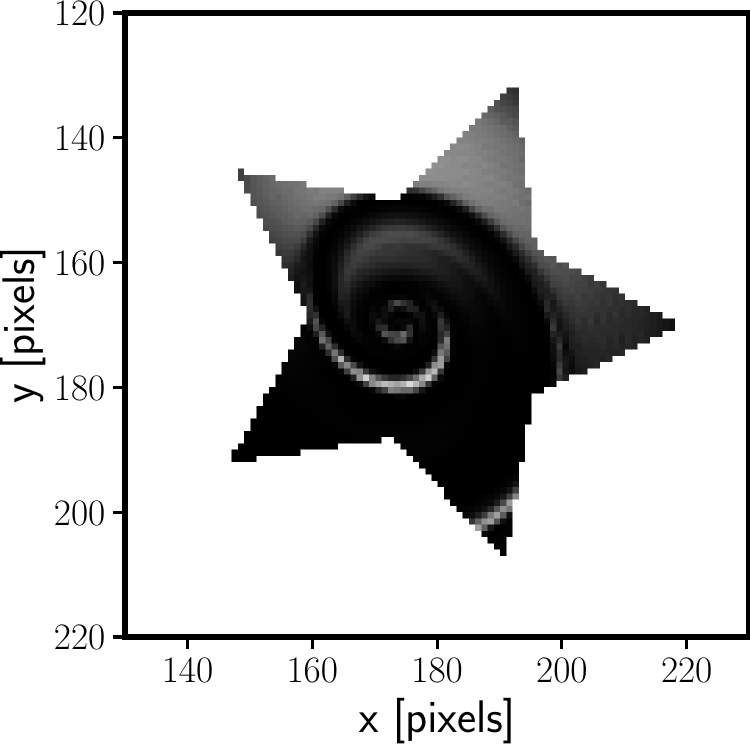} &
    \includegraphics[width=\texW]{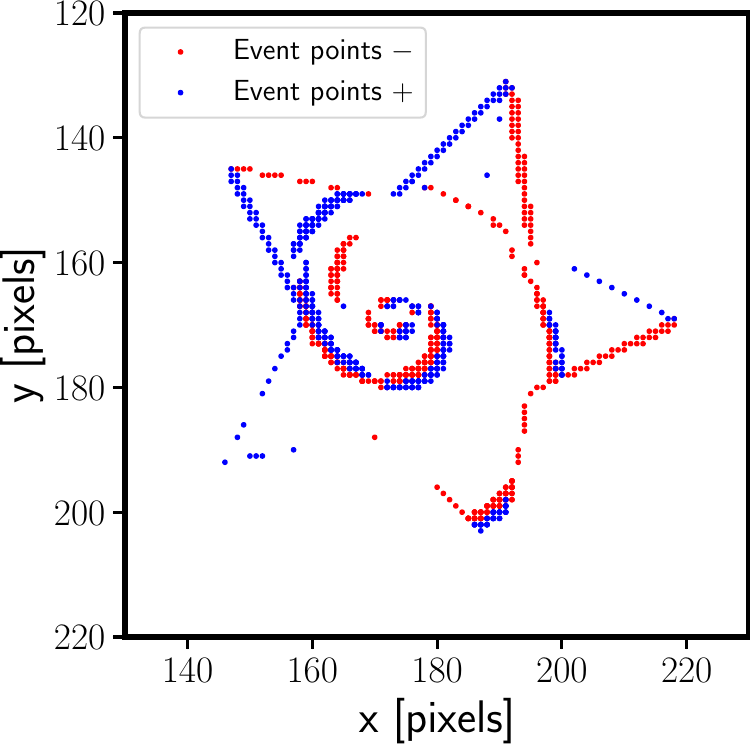} \\
  \end{tabular}
  \caption{Toy dataset with different foreground textures from DTD~\cite{cimpoi14describing} (part~1 of~2).
    \textbf{Left}: original texture image applied to the star shape.
    \textbf{Middle}: rendered intensity frame.
    \textbf{Right}: generated event polarity scatter (blue~$=$~positive, red~$=$~negative).}
  \label{fig:supp:texture_visualization_1}
\end{figure*}

\begin{figure*}[h]
  \centering
  \begin{tabular}{@{}c@{\hskip 4pt}c@{\hskip 4pt}c@{}}
    \textbf{Original Texture} & 
    \parbox[b][0.1cm][c]{\texW}{\centering\textbf{Intensity Frame \\at $t{=}203$~ms}} & 
    \parbox[b][0.1cm][c]{\texW}{\centering\textbf{Events \\at $t{=}200$--$205$~ms}} \\[4pt]
    \parbox[b][\texW][c]{\texW}{\centering\includegraphics[width=\texW]{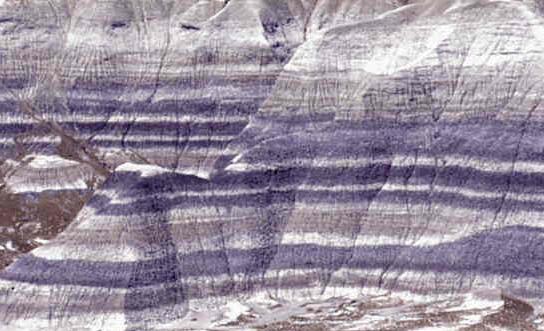}\\[2pt]{\small stratified}} &
    \includegraphics[width=\texW]{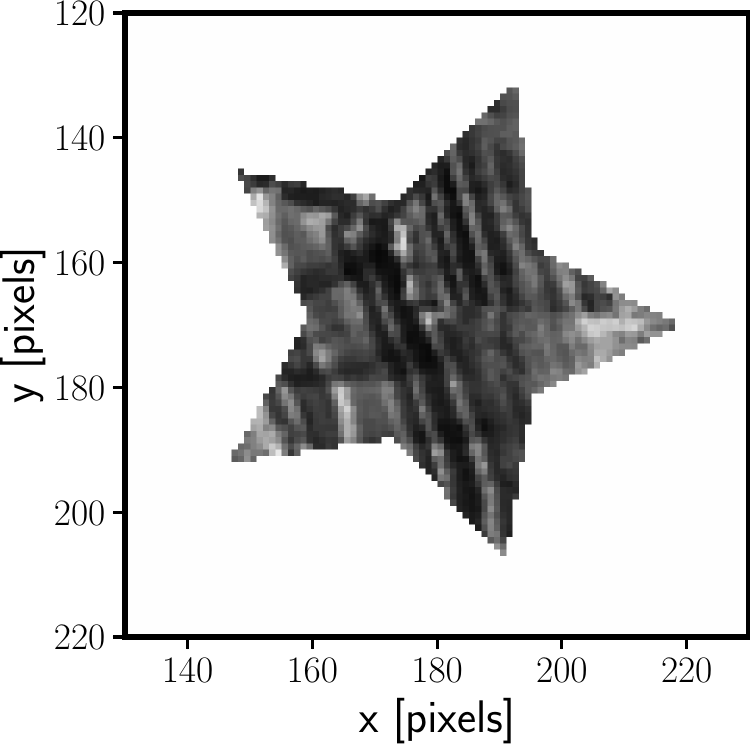} &
    \includegraphics[width=\texW]{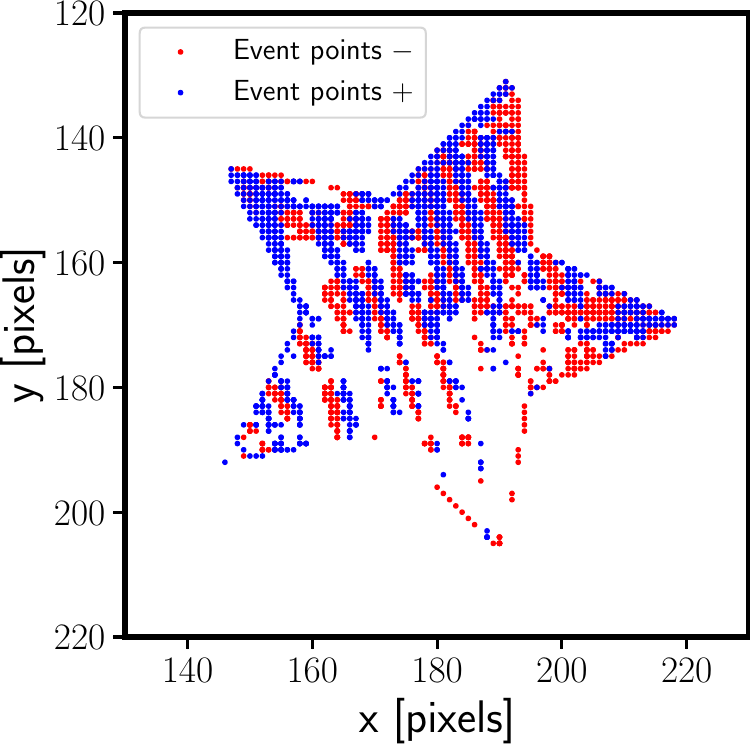} \\[4pt]
    \parbox[b][\texW][c]{\texW}{\centering\includegraphics[height=0.25\textwidth]{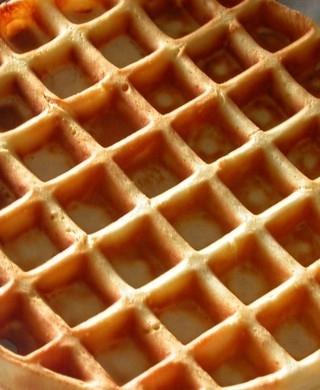}\\[2pt]{\small waffled}} &
    \includegraphics[width=\texW]{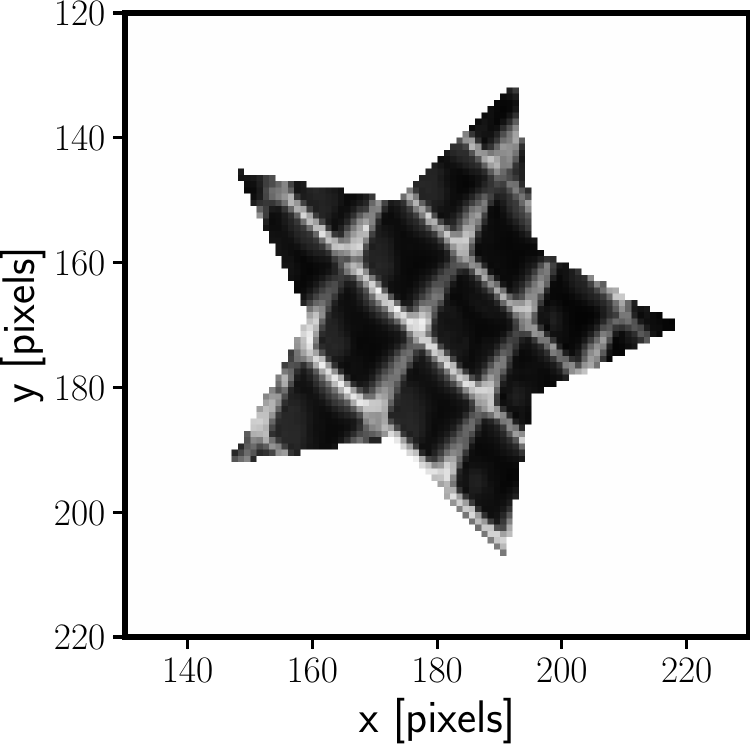} &
    \includegraphics[width=\texW]{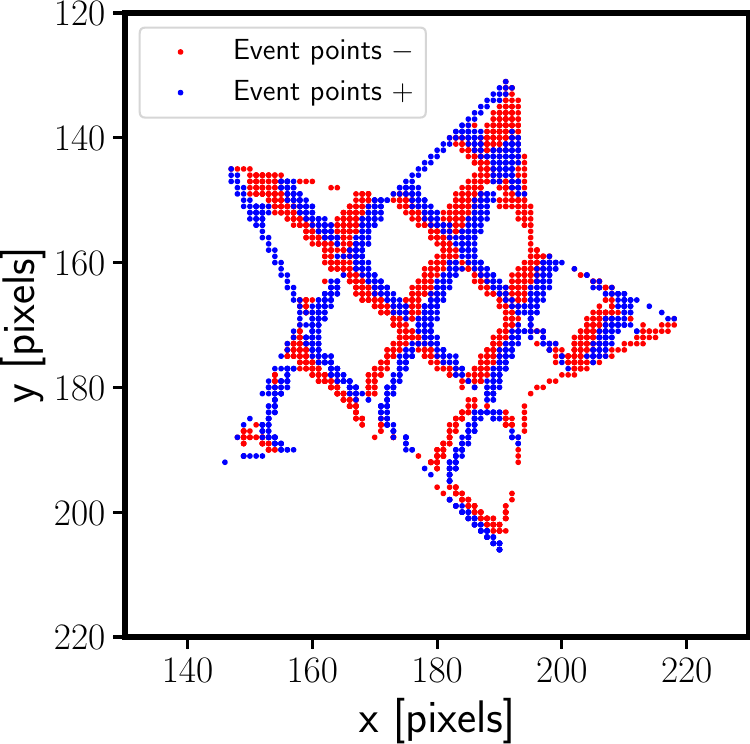} \\[4pt]
    \parbox[b][\texW][c]{\texW}{\centering\includegraphics[height=0.25\textwidth]{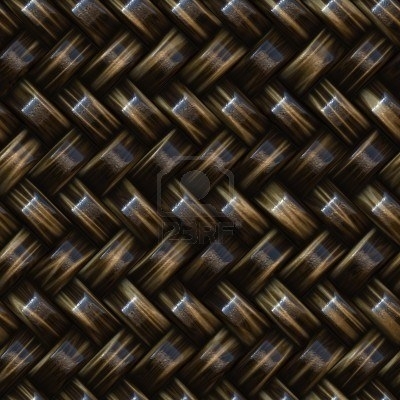}\\[2pt]{\small woven}} &
    \includegraphics[width=\texW]{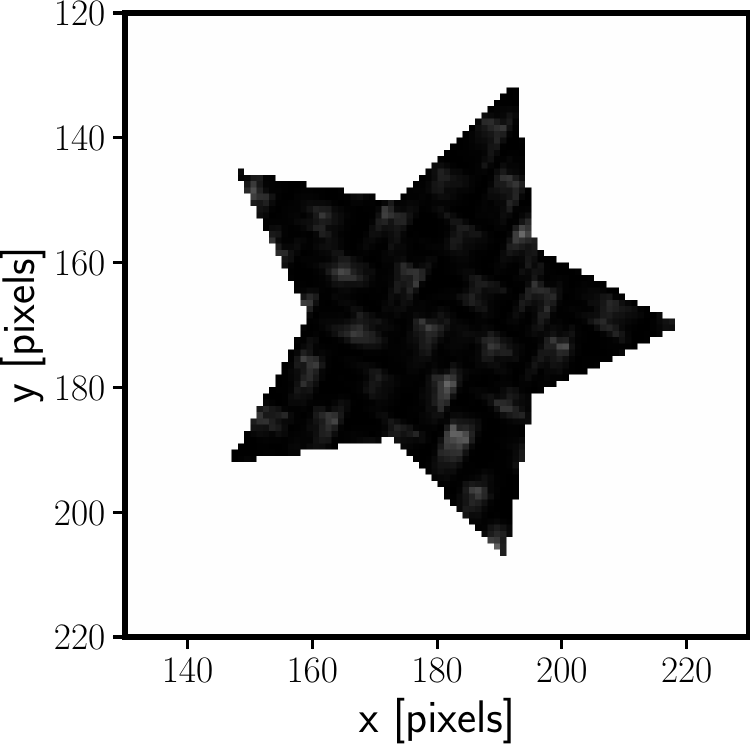} &
    \includegraphics[width=\texW]{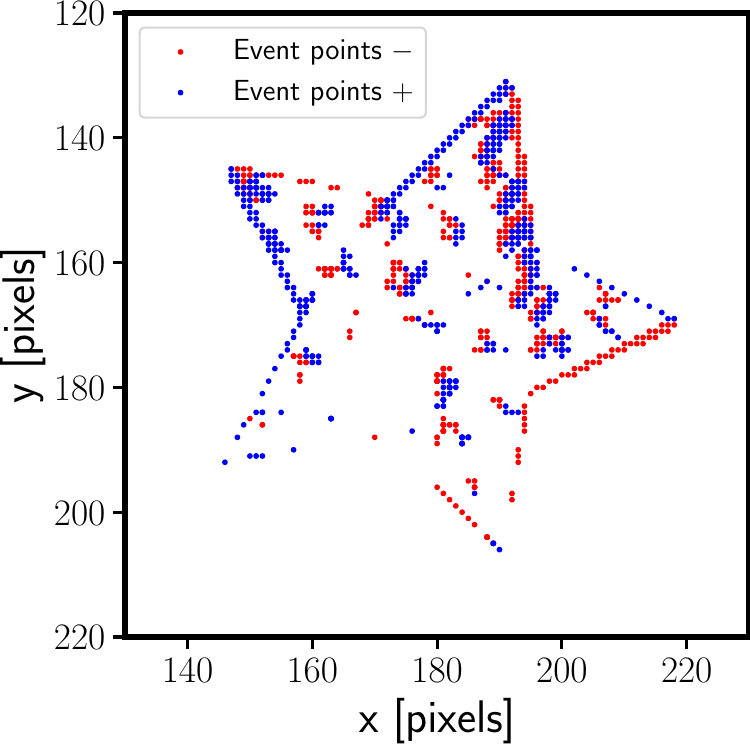} \\
  \end{tabular}
  \caption{Toy dataset with different foreground textures from DTD~\cite{cimpoi14describing} (part~2 of~2).
    \textbf{Left}: original texture image applied to the star shape.
    \textbf{Middle}: rendered intensity frame.
    \textbf{Right}: generated event polarity scatter (blue~$=$~positive, red~$=$~negative).}
  \label{fig:supp:texture_visualization_2}
\end{figure*}

\begin{figure}[h]
  \centering
  \begin{subfigure}[t]{0.49\textwidth}
    \centering
    \includegraphics[width=\linewidth]{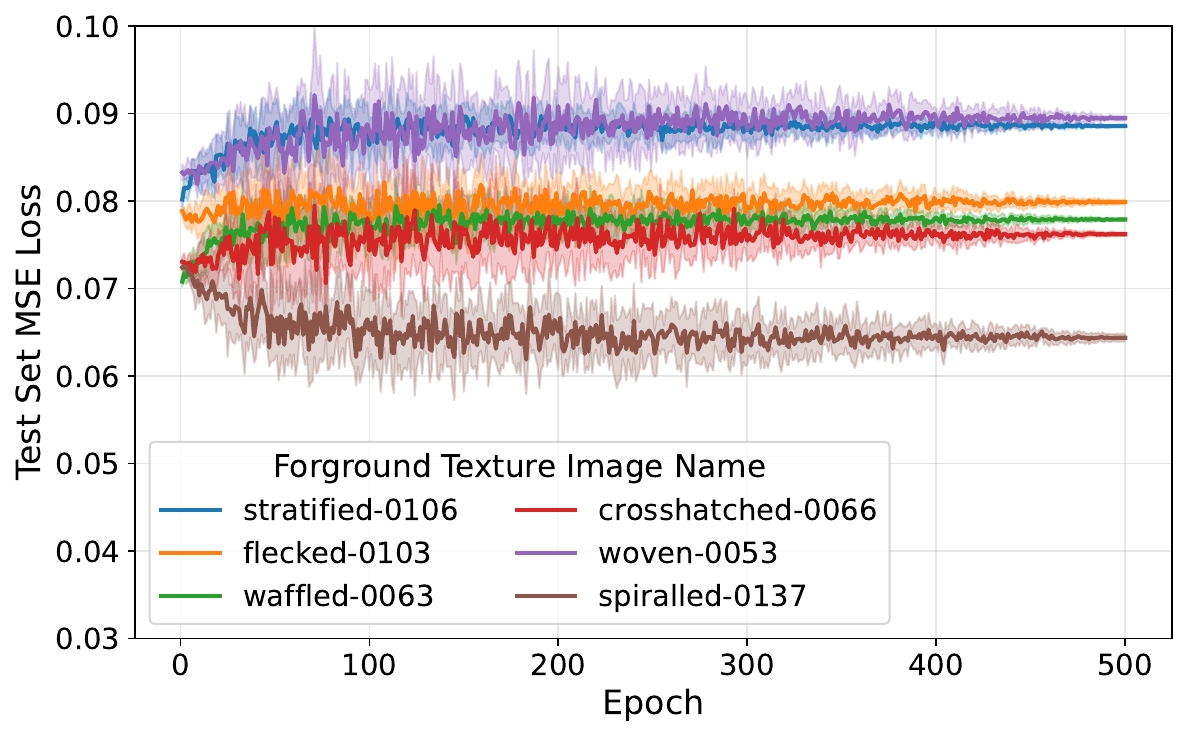}
    \caption{Baseline}
    \label{fig:toy:texture original}
  \end{subfigure}
  \begin{subfigure}[t]{0.49\textwidth}
    \centering
    \includegraphics[width=\linewidth]{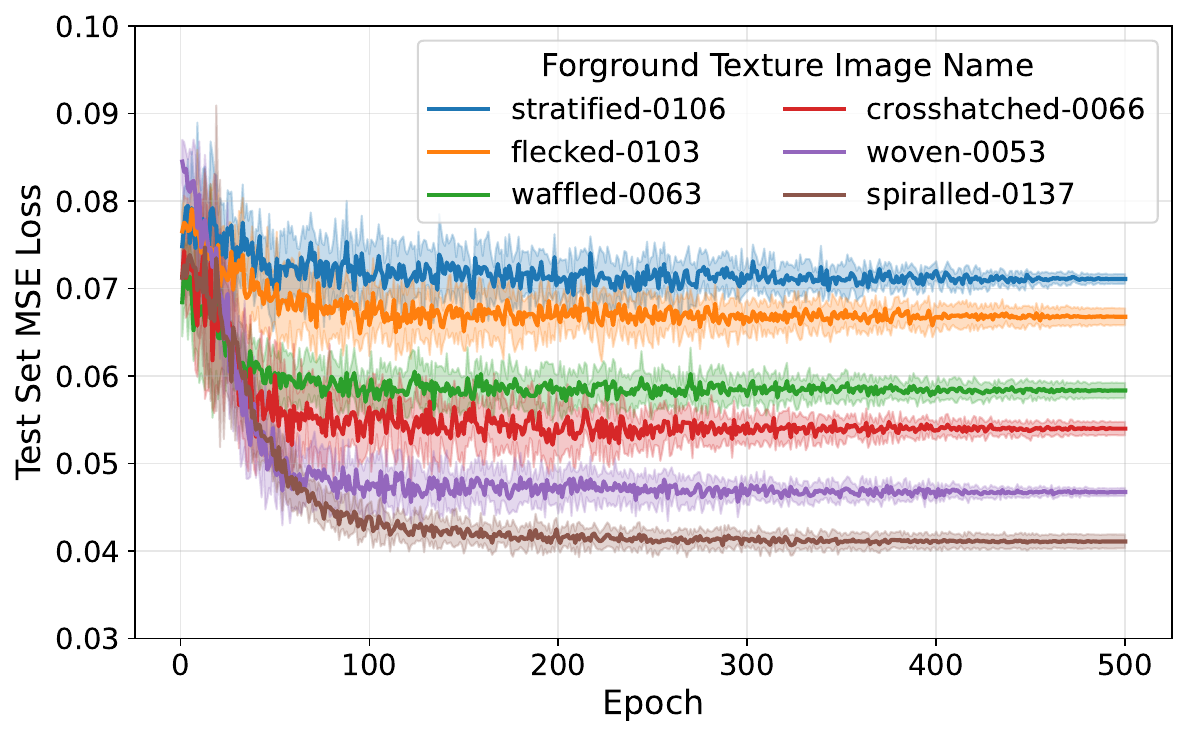}
    \caption{With Eigenvalues + Density Values}
    \label{fig:toy:texture both}
  \end{subfigure} 
\caption{Effect of adding various textures to the moving foreground object. With texture, the combined feature model still learns to estimate optical flow, while the baseline breaks down as the dataset complexity increases.}
  \label{fig:toy:texture}
\end{figure}

\begin{figure}[h]
  \centering
  \begin{subfigure}[t]{0.49\textwidth}
    \centering
    \includegraphics[width=\linewidth]{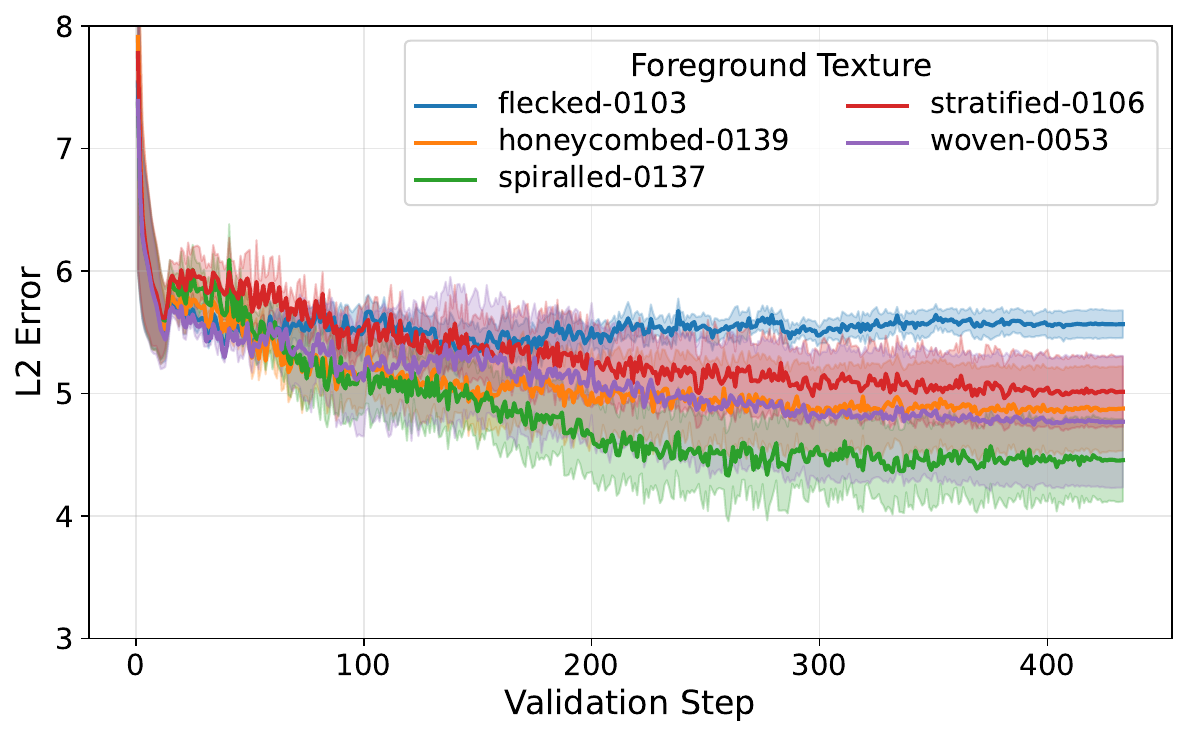}
    \caption{\Originalcase{}}
    \label{fig:toy:tinyidnet:texture original}
  \end{subfigure}
  \begin{subfigure}[t]{0.49\textwidth}
    \centering
    \includegraphics[width=\linewidth]{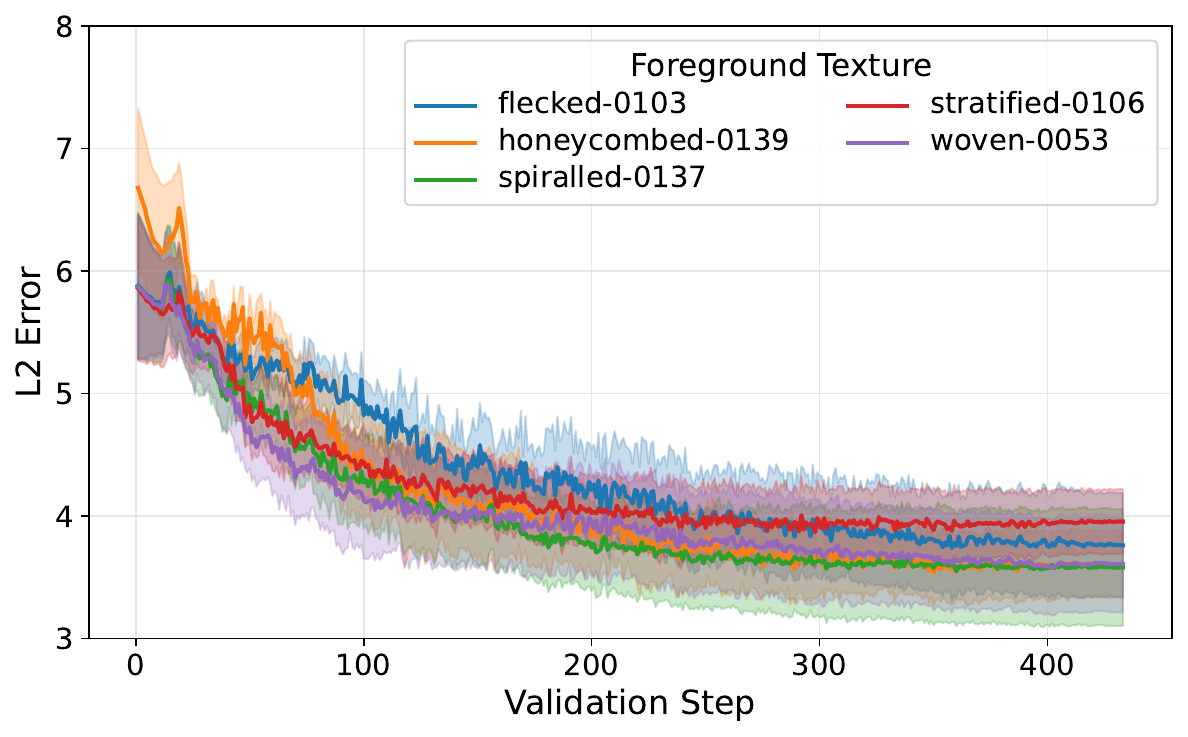}
    \caption{with \Eigvalues{} + \Filtervalues{}}
    \label{fig:toy:tinyidnet:texture both}
  \end{subfigure} 
\caption{Effect of adding texture to the moving foreground object using \emph{Tiny IDNet} (\texttt{rnn\_input\_dim}=4, \texttt{hidden\_dim}=4). Compared to the \originalcase{}, extending with \Eigvalues{} and \Filtervalues{} achieves lower MSE error at test time.}
  \label{fig:toy:tinyidnet:texture}
\end{figure}

\begin{figure*}[h]
  \centering
  \begin{subfigure}[t]{0.32\textwidth}
    \centering
    \includegraphics[width=\linewidth]{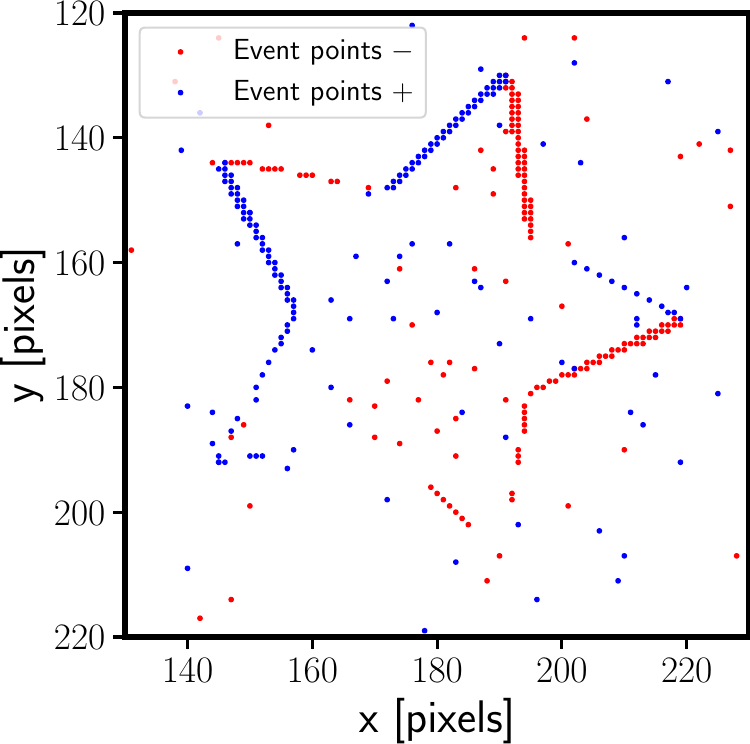}
    \caption{\small 5\,Hz}
  \end{subfigure}
  \hfill
  \begin{subfigure}[t]{0.32\textwidth}
    \centering
    \includegraphics[width=\linewidth]{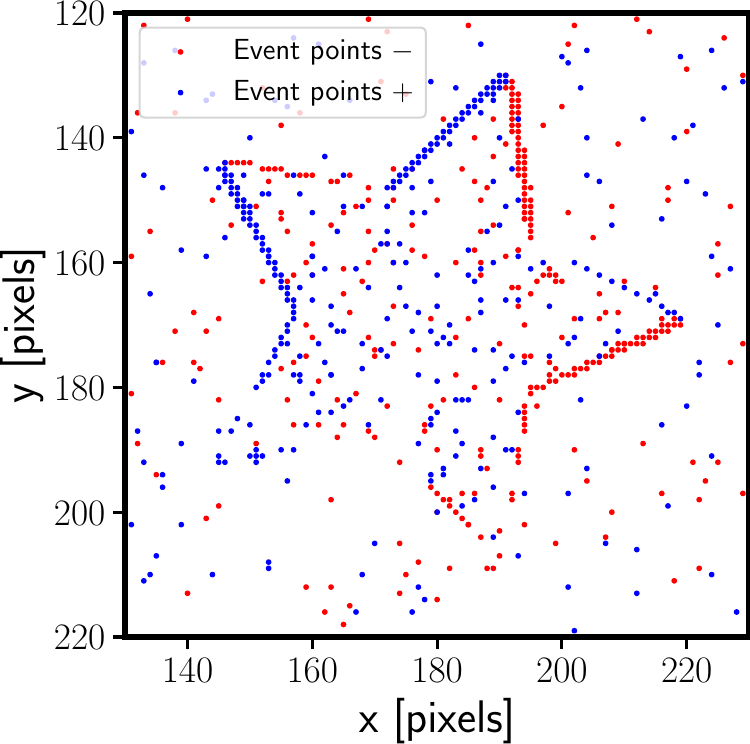}
    \caption{\small 20\,Hz}
  \end{subfigure}
  \hfill
  \begin{subfigure}[t]{0.32\textwidth}
    \centering
    \includegraphics[width=\linewidth]{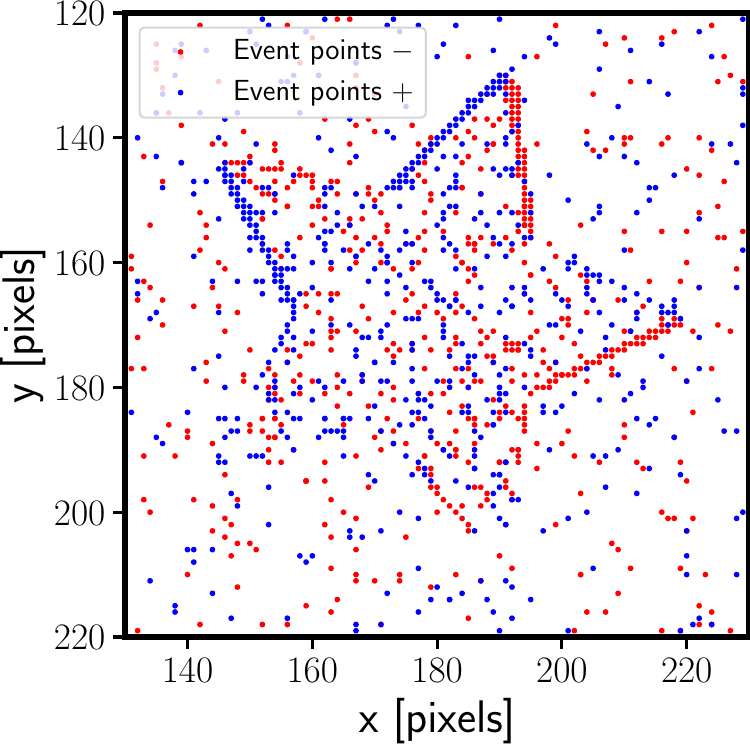}
    \caption{\small 50\,Hz}
  \end{subfigure}
  \caption{Toy dataset with different levels of shot noise generated by V2E~\cite{hu2021v2e}.
    Each subfigure shows the generated event polarity scatter at a different noise frequency.
    Higher shot-noise frequency produces more spurious events and visibly stronger noise contamination.}
  \label{fig:supp:shot_noise_visualization}
\end{figure*}

\begin{table*}[h]
\centering
\caption{Performance under shot noise events. Comparing \Originalcase{} and \bothshortvalues{} features at 5 Hz and 50 Hz noise frequencies using the \toymodel{} (hidden dim 128, $K=50$). Best values in \textbf{bold}.}
\label{tab:toy:shot_noise_comparison}
\begin{adjustbox}{width=\textwidth}%

\begin{tabular}{@{}lcccccccc@{}}
\toprule
\multirow{2}{*}{Feature Type}       & \multicolumn{4}{c}{5 Hz} & \multicolumn{4}{c}{50 Hz} \\ \cmidrule(lr){2-5}\cmidrule(l){6-9} 
                                    & L1 {\small $(\times 10^{-1})$} & MSE {\small $(\times 10^{-2})$} & EPE {\small $(\times 10^{-1})$} & AE & L1 {\small $(\times 10^{-1})$} & MSE {\small $(\times 10^{-2})$} & EPE {\small $(\times 10^{-1})$} & AE \\ \midrule
\Originalcase{} & $1.452 \smallpm{0.015}$ & $3.610 \smallpm{0.080}$ & $2.275 \smallpm{0.023}$ & $42.6 \smallpm{0.5}$ & $2.031 \smallpm{0.009}$ & $6.050 \smallpm{0.050}$ & $3.149 \smallpm{0.014}$ & $67.1 \smallpm{0.5}$ \\
\bothshortvalues{} & $\mathbf{0.761} \smallpm{0.009}$ & $\mathbf{1.040} \smallpm{0.020}$ & $\mathbf{1.189} \smallpm{0.013}$ & $\mathbf{17.8} \smallpm{0.3}$ & $\mathbf{1.069} \smallpm{0.017}$ & $\mathbf{1.960} \smallpm{0.050}$ & $\mathbf{1.674} \smallpm{0.026}$ & $\mathbf{25.9} \smallpm{0.5}$ \\
\bottomrule
\end{tabular}
\end{adjustbox}
\end{table*}

\begin{figure}[h]
  \centering
    \includegraphics[width=0.6\linewidth]{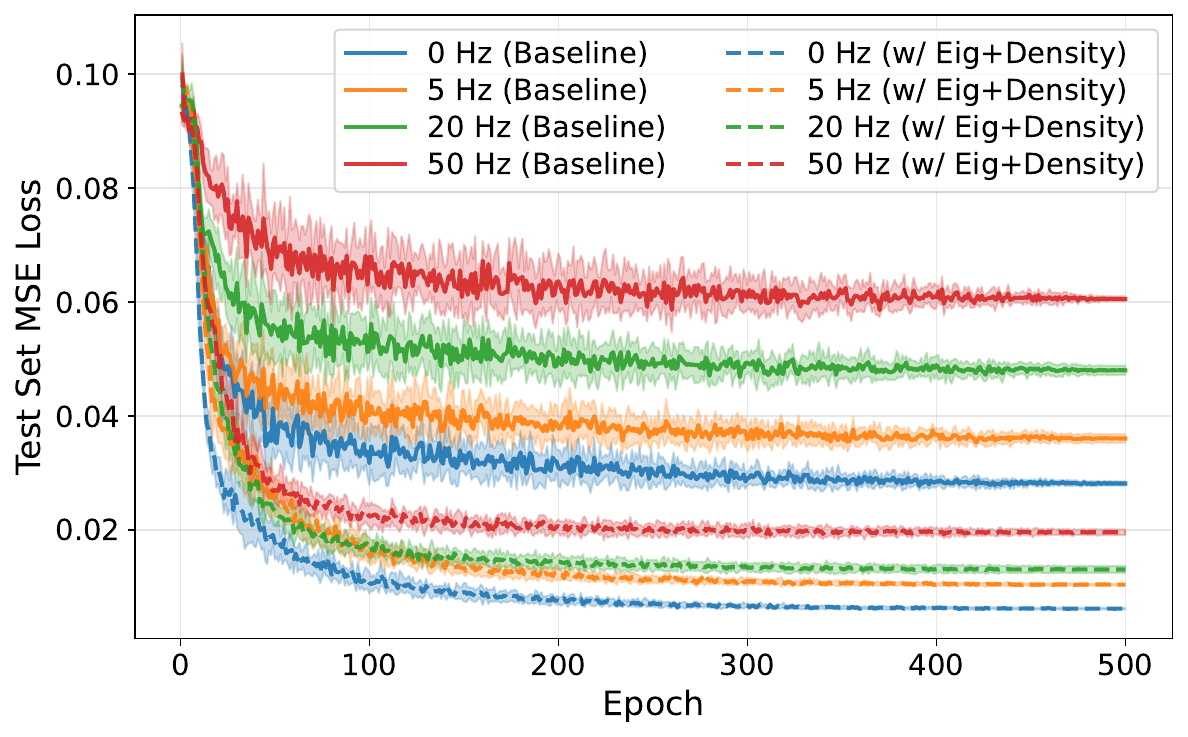}
\caption{Shot noise robustness across different noise frequencies (higher frequency means noisier events). The combined feature model (dotted) remains robust and consistently outperforms the baseline (solid).}
\label{fig:toy:shot noise}
\end{figure}


\else
\begin{abstract}
  Event cameras capture intensity changes asynchronously with high temporal resolution, requiring novel preprocessing methods for downstream tasks.
  Unlike static intensity snapshots, event data inherently encode information about scene dynamics and object motion, meaning that features derived from events can exhibit behaviors with no direct analogue in frame-based vision.
  In this paper, we analyze two features used in event-based corner detection---the eigenvalues of the structure tensor and the spatiotemporal density values---and show that they are \emph{motion cues}.
  We hypothesize that these features, combined with local geometric information, can enhance motion estimation tasks.
To validate this, we first theoretically analyze how the eigenvalues of the structure tensor at moving corner points relate to the direction of motion.
  We then design controlled experiments on a synthetic dataset, confirming that extending local geometric features with eigenvalues and density values provides complementary motion information and is robust to texture and shot noise.
  Finally, we integrate the proposed features into a state-of-the-art event-based optical flow network and evaluate on the real-world DSEC benchmark, where the added features consistently improve accuracy, with the largest gains in data-scarce scenarios and for lower-capacity models.
The code for this paper can be found at: \href{https://github.com/hesamaraghi/static-in-frames-dynamic-in-events}{https://github.com/hesamaraghi/static-in-frames-dynamic-in-events}.

\end{abstract}
\section{Introduction}
\label{sec:intro}

Event cameras differ fundamentally from conventional frame-based cameras.
Instead of capturing absolute intensity values at fixed time intervals, they asynchronously report per-pixel intensity \emph{changes} with microsecond temporal resolution~\cite{gallego_event-based_2022}.
While a frame-based camera captures instantaneous snapshots, an event camera continuously measures temporal derivatives of the visual signal.
This sensing paradigm means that event data inherently encodes information about \emph{motion} in ways that static frames do not.
Because events are triggered by changes in intensity, their spatiotemporal distribution implicitly reflects the dynamics of the scene and the motion of objects. 
Consequently, features derived from event data can exhibit behaviors that have no direct analogue in frame-based vision, where features are typically computed from static intensity snapshots and lack explicit motion encoding.
Thus, if representations of the event features already contain motion information, they can help as motion cues for downstream estimation.

\begin{figure}[t]
    \centering
    \begin{subfigure}[b]{0.24\columnwidth}
        \centering
        \includegraphics[width=\linewidth]{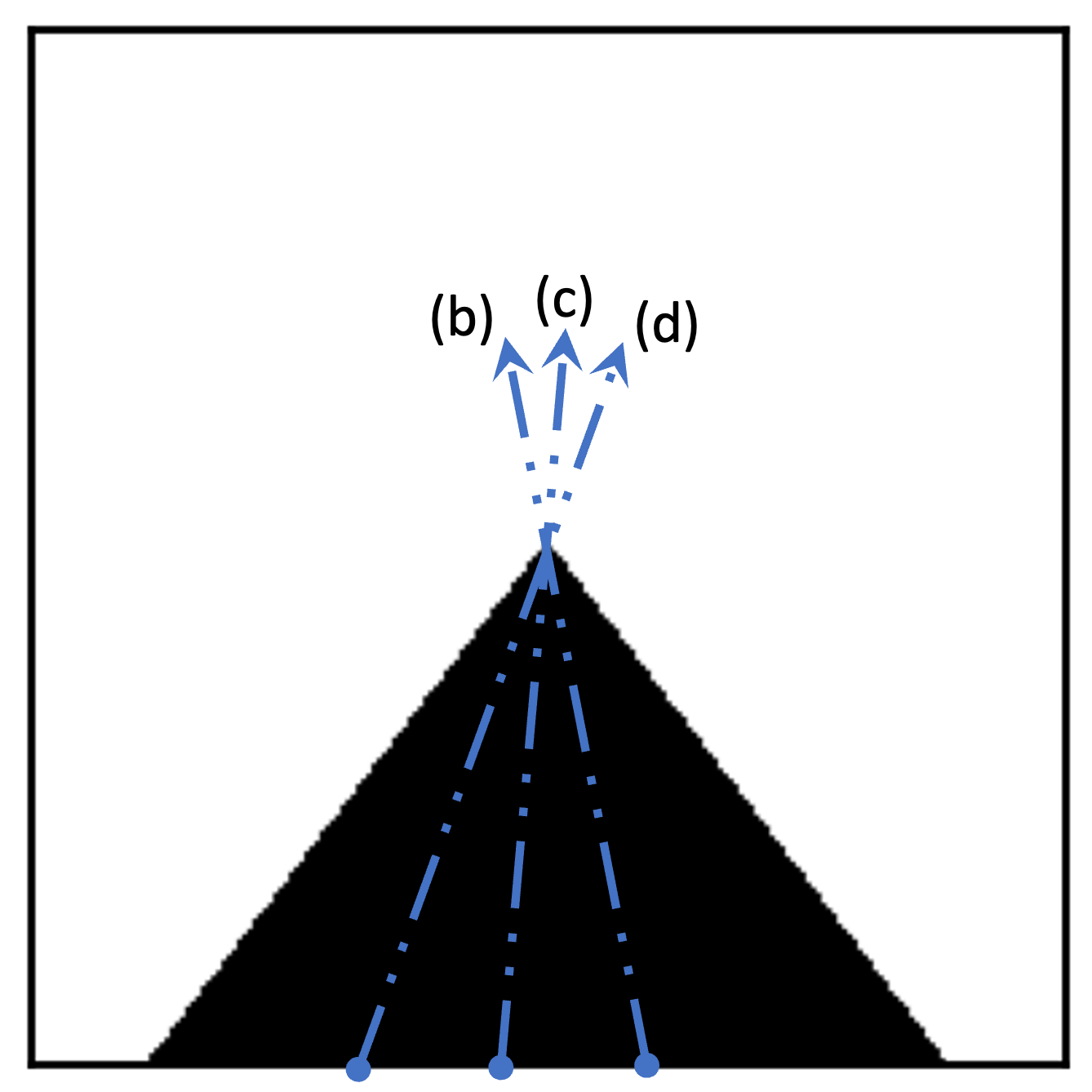}
        \caption{Triangle motion}
        \label{fig:fig1:triangle shape}
    \end{subfigure}
    \hfill
    \begin{subfigure}[b]{0.24\columnwidth}
        \centering
        \includegraphics[width=\linewidth]{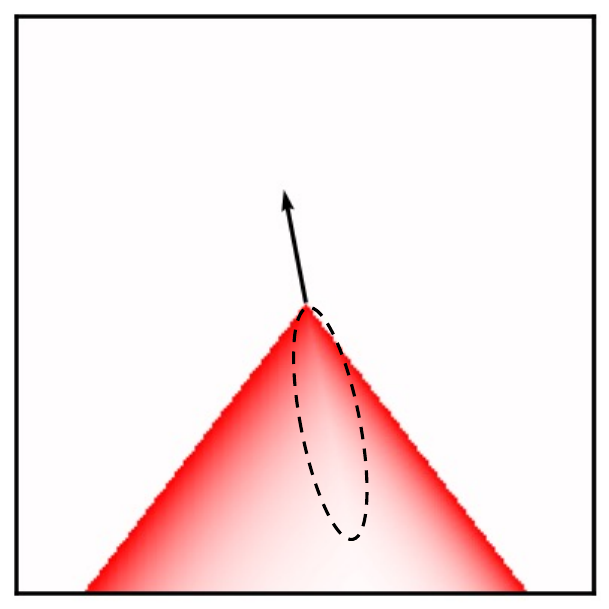}
        \caption{motion in $-11^\circ$}
        \label{fig:fig1:dir -11}
    \end{subfigure}
    \hfill
    \begin{subfigure}[b]{0.24\columnwidth}
        \centering
        \includegraphics[width=\linewidth]{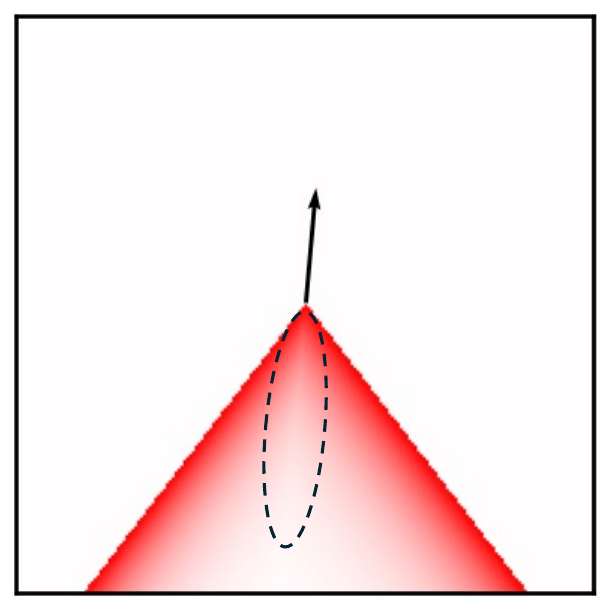}
        \caption{motion in $5^\circ$}
        \label{fig:fig1:dir 5}
    \end{subfigure}
    \hfill
    \begin{subfigure}[b]{0.24\columnwidth}
        \centering
        \includegraphics[width=\linewidth]{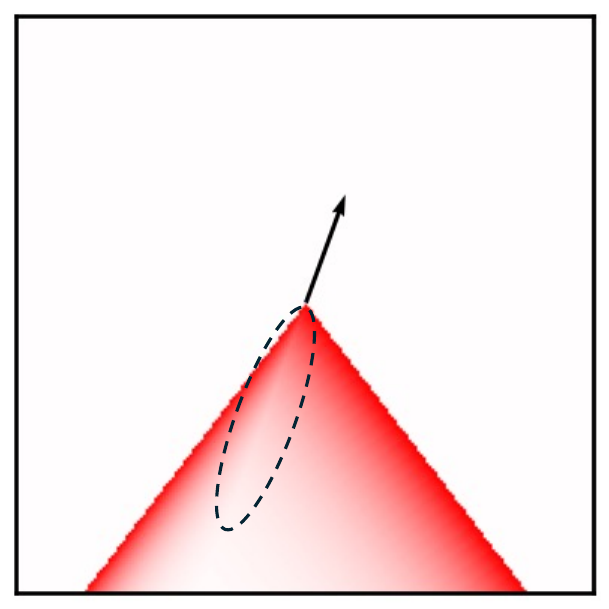}
        \caption{motion in $20^\circ$}
        \label{fig:fig1:dir 20}
    \end{subfigure}
    \caption{Motion direction encoded in event-based density features. (a) A triangular shape moving in different directions. (b--d) Density visualizations for three different motion directions: the oriented traces in the density maps (dashed lines) align with the motion direction. Red indicates higher density of recent events.}
    \label{fig:fig1 triangle motion direction}
\end{figure}


\Cref{fig:fig1 triangle motion direction} presents a synthetic experiment illustrating how the spatiotemporal density of events encodes motion direction.
A triangular shape translates in three directions, with angles of $-11^\circ$, $5^\circ$, and $20^\circ$ (\cref{fig:fig1:triangle shape}).
For each direction, we generate an event sequence and compute the density of recent events within a local spatiotemporal neighborhood.
In the visualization, red indicates higher density of recent events.
The resulting density representations exhibit clear, continuous traces whose orientation matches the underlying motion.
This shows that spatiotemporal density values naturally encode directional motion information, a property that follows from the asynchronous event-sensing mechanism.

In this paper, we analyze the features involved in event corner detection, specifically the eigenvalues of the structure tensor and the spatiotemporal density values.
We consider the Harris corner detector~\cite{Harris1988} approach for event cameras~\cite{vasco_fast_2016, li_fa-harris_2019, glover_luvharris_2022}.
The common procedure for finding corners first generates a local grid representation around the event, then computes the structure tensor from the local gradients of the grid representation.
The Harris eigenvalues of the structure tensor serve as indicators of corner scores.
To obtain the local grid representation, we spatiotemporally filter the events to compute a density-based representation, and we compute the eigenvalues of the structure tensor from the gradients of this local grid.
For event input, we discover that the \emph{\eigvalues{}} and \emph{\filtervalues{}} features are `motion cues', meaning they carry information about scene dynamics and object motion.
We hypothesize that these features, along with local geometric features, can aid motion estimation tasks.
\nergis{This is not entirely true? e.g. the eigenvalues are computed from spatiotemporal density, which includes at least recent history of event, e.g. it's not static in time...? I'd remove the previous sentence.}
To validate our hypothesis, we conduct a theoretical analysis, controlled synthetic experiments, and real-world evaluation. 

Our main contributions are summarized as follows:
\begin{itemize}
\item We establish a theoretical connection between the \eigvalues{} of the structure tensor and motion direction in event cameras. In particular, we show how \eigvalues{}, when combined with local geometric information, can be used to recover the motion direction of translating corners.

\item We design controlled experiments on synthetic datasets to validate that extending local geometric features with \eigvalues{} and \filtervalues{} improves optical flow estimation. 
We demonstrate that these improvements stem from informative and complementary motion cues by showing that a network can learn the mapping from these features to optical flow.
We further show that the proposed extending features are robust under challenging conditions, including textured foregrounds and shot noise.

\item We validate our findings on the real-world DSEC benchmark~\cite{gehrig_dsec_2021} using IDNet~\cite{wu_lightweight_2024}, showing consistent improvements in optical flow estimation. Notably, larger gains are observed in data-scarce scenarios and for lower-capacity models.

\end{itemize}
\section{Related works}
\label{sec:related_work}

\subsection{Event representations}
\label{sec:related_work subsec:event representations}
Converting the asynchronous event stream into a suitable representation is critical for downstream tasks.
Common representations include event count~\cite{maqueda_event-based_2018}, time surfaces that record the most recent timestamp at each pixel~\cite{lagorce_hots_2017,sironi_hats_2018}, voxel grids ~\cite{zhu_unsupervised_2019}, and TORE volumes that store raw spike timing information~\cite{baldwin_time-ordered_2021}.
Voxel grid \cite{zhu_unsupervised_2019} is a discretized event volume that accumulates events in temporal bins, which is widely used for event-based optical flow estimation.
EventPoint~\cite{huang_eventpoint_2023} proposed Tencode, converting the event stream into the 3-channel structure similar to RGB input.
Most of these representations are designed empirically for aggregating spatiotemporal information for general-purpose processing.
In contrast, our work focuses on features that encode complementary motion information, accompanied by theoretical analysis on the explicit connection between these features and motion direction.

Another line of works consider hybrid representations, fusing event data with other modalities.
Gehrig~\etal~\cite{gehrig_dense_2024} combine frames and events for dense continuous-time optical flow by using separate encoders for each modality.
Wan~\etal~\cite{wan_learning_2022} learn dense optical flow from events by fusing event and image information.
Zhou~\etal~\cite{zhou_bridge_2025} bridge frame and event modalities through a common spatiotemporal fusion strategy for high-dynamic scene optical flow.
Beyond bimodal approaches, RPEFlow~\cite{wan_rpeflow_2024} fuses RGB, LiDAR point cloud, and event data for joint optical flow and scene flow estimation.
Sun~\etal~\cite{sun_event-based_2022} apply cross-modal attention to fuse events and frames for motion deblurring.
These methods typically employ separate encoders for each modality and rely on feature-level attention or cross-attention mechanisms to integrate information across modalities.
While our work shares a similar philosophy of extending the representation with complementary information, it is more lightweight.
Its plug-and-play design makes it easy to integrate into existing architectures without the need for separate encoders or complex fusion strategies.

\subsection{Event-based optical flow estimation}
\label{sec:related_work subsec:optical flow estimation}
Optical flow estimation is a fundamental task in computer vision that serves as a natural benchmark for assessing how well a method captures motion cues.
Existing methods for event-based optical flow estimation can be divided into supervised or unsupervised approaches.
Self-supervised methods~\cite{zhu_unsupervised_2019, hagenaars_self-supervised_2021, paredes-valles_taming_2023,hamann_motion-prior_2024} learn optical flow typically using photometric or contrast maximization~\cite{gallego_unifying_2018,gallego_focus_2019,shiba_secrets_2022} losses.
Supervised methods, on the other hand, generally achieve higher accuracy.
E-RAFT~\cite{gehrig_e-raft_2021} builds upon the idea of RAFT~\cite{teed_raft_2020} in using correlation volumes for event-based optical flow estimation.
TMA~\cite{liu_tma_2023} exploits temporal continuity of events through a temporal motion aggregation strategy.
To reduce the computational complexity, IDNet~\cite{wu_lightweight_2024} avoids the computationally expensive correlation volume component and instead directly estimates flow via iterative deblurring, achieving a favorable balance between accuracy and computational efficiency.
These methods typically use voxel grids as spatiotemporal representations of events. In this work, we extend this representation with complementary motion features and show experimentally that they improve optical flow estimation.
\section{Method}
\label{sec:method}

\subsection{Feature extraction}
\label{sec:method subsec:feature extraction}

An event camera produces a stream of asynchronous events $\{e_i\}_{i=1}^{N}$, where each event ${e_i = (x_i, y_i, t_i, p_i)}$ encodes a pixel location $(x_i, y_i)$, a timestamp $t_i$, and a polarity $p_i \in \{-1, +1\}$ indicating brightness increase or decrease.
From this event stream, we follow the approach in \cite{araghi2025making} to construct an image $I$ of \emph{\filtervalues{}s} that captures the causal spatiotemporal density of events.
Specifically, as each event $e_i$ arrives, we update the \filtervalues{} at the $w_d \times w_d$ spatial neighborhood of $e_i$ by accumulating contributions from past events, weighted by polarity, spatial proximity, and temporal recency~\cite{araghi2025making}, i.e.,
\begin{equation}
I(\vec{r}_i) = \sum_{j \leq i} p_j \cdot g(\vec{r}_i - \vec{r}_j) \exp\left(\frac{t_j - t_i}{\tau}\right),
\label{eq:filter_value}
\end{equation}
where $\vec{r}_i = (x_i, y_i)$ denotes the spatial position, $g(\cdot)$ is a Gaussian spatial kernel of size $w_d \times w_d$ with standard deviation $\sigma = w_d/5$, and $\tau$ is the temporal decay constant.
The exponential temporal kernel enables recursive computation as the \filtervalues{}s are updated efficiently as each event arrives without buffering past events.
The result is a 2D image $I$ where each pixel value reflects the recent event density in its neighborhood.

From the density image $I$, we compute spatial gradients $\vec{\nabla I} = [I_x, I_y]^T$ using Sobel operators, where $I_x$ and $I_y$ denote the horizontal and vertical derivatives of $I$.
We then compute the structure tensor $S_w$ at the event location, following the Harris corner detector formulation~\cite{Harris1988}:
\begin{equation}
S_w(\vec{r}_i) = w * \begin{bmatrix} I_x^2 & I_x I_y \\ I_x I_y & I_y^2 \end{bmatrix}= w * (\vec{\nabla I} \vec{\nabla I}^T),
\label{eq:structure_tensor_feature}
\end{equation}
where $w$ is a window function and $*$ denotes convolution.
The structure tensor encodes the distribution of gradient orientations in a local neighborhood; its \eigvalues{} $\lambda_1$ and $\lambda_2$ are computed in closed form as:
\begin{equation}
\lambda_{1,2} = \frac{1}{2}\left( \mathrm{tr}(S_w) \pm \sqrt{(\mathrm{tr}(S_w))^2 - 4\det(S_w)} \right).
\label{eq:eigenvalues}
\end{equation}

The extracted per-event features, the density value $I(\vec{r}_i)$ and the \eigvalues{} pair $(\lambda_1, \lambda_2)$, are used to extend baseline representations.
In \cref{sec:method subsec:triangle theory}, we analyze how these eigenvalues relate to motion cues.

\subsection{From Eigenvalues to Motion Cues}
\label{sec:method subsec:triangle theory}

\begin{figure}[t]
    \centering
    \includegraphics[width=0.5\linewidth]{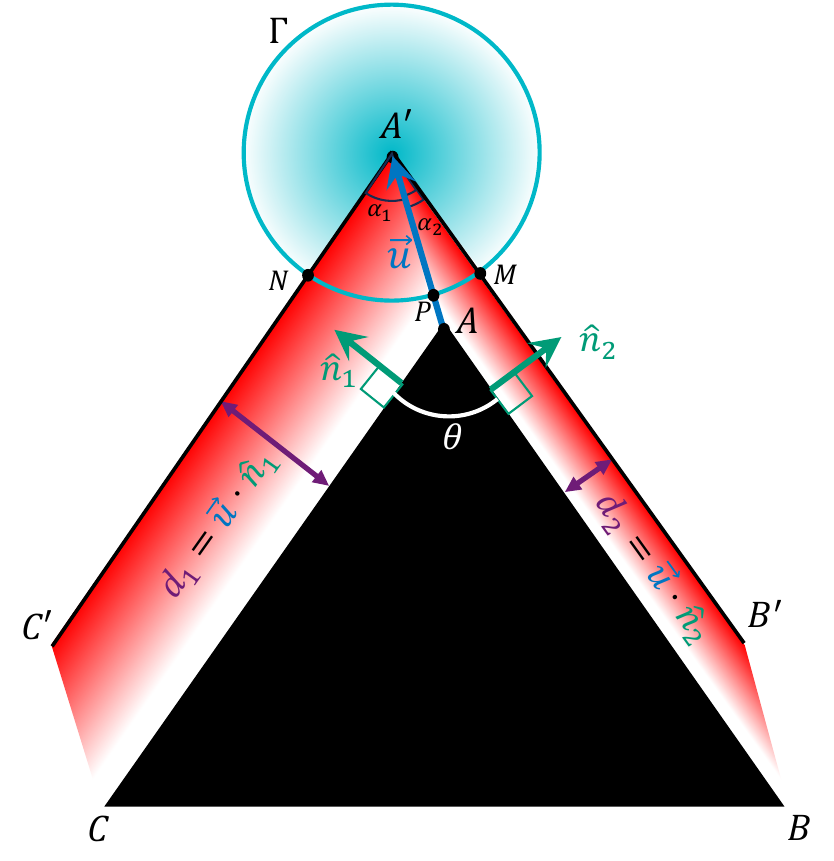}
    \caption{%
Establishing the relationship between local geometry, eigenvalues and motion direction in an example of a triangle translation. The triangle $\triangle CAB$ translates to $\triangle C'A'B'$ along vector $\vec{u}$. The red shading encodes event recency, with more recent events near $\triangle C'A'B'$. Unit normals $\vec{{\hat n}}_1$ and $\vec{{\hat n}}_2$ at vertex $A$ define the local geometry. \nergis{Can we mention here something about how we treat polarity? \eg that the triangle is white/black and the background is black/white and all the events are a certain polarity? etc.}
    }
    \label{fig:theory_trianlge}
\end{figure}

Here, we theoretically establish the connection between \eigvalues{} and motion direction using a simple translation example of a triangle, as illustrated in \cref{fig:theory_trianlge}. 
The triangle $\triangle CAB$ moves to $\triangle C'A'B'$  with displacement vector $\vec{u}$, without any rotation or deformation. 
The red shading in the figure indicates recent event density after the motion is complete: newer events (near $\triangle C'A'B'$) appear more red, while older ones fade to white. 
\nergis{I wonder if reviewers would ask here, if temporal recency is all that matters, why not use a time surface? I think one of the main points is that, depending on the direction of movement, one side of the triangle emits more events (not just more recent events), than the other, so the density-based filter has higher values... is that true?}\hesam{The density-based is applying a exponential decay on the timestamps of the events, so it is very similar to time surface in this example. But the difference can be seen for example when we have an object with texture moving. In this case, density-based can capture the distribution (density) of the timestamps and it can indicate the speed of movement.} \nergis{Good answer! :) Maybe density is also important for different/more complex shapes, e.g. for the star shape the density will be different for the convex corners vs. concave corners, even if both have very recent events...}

Our goal is to study the hypothesis that \eigvalues{}, combined with local geometry information, delivers motion cues, by examining the motion of vertex $A$.
As local geometry, we assume we know the shape of the triangle, with $A$ a convex corner with angle $\theta$.
More specifically, we assume we have the unit normal vectors of $\vec{\hat n}_1$ and $\vec{\hat n}_2$ for edges $AC$ and $AB$, and we construct matrix
$V=\begin{bmatrix}\vec{\hat n}_1, \vec{\hat n}_2\end{bmatrix}$.
Matrix $V$ thus represents the local geometric structure.
Let ${\vec{d} = V^T \vec{u}}$, whose entries correspond to the projections of $\vec{u}$ onto each normal, \ie $\vec{d}=[d_1, d_2]^T \triangleq [\vec{u}\!\cdot\!\vec{\hat n}_1, \, \vec{u}\!\cdot\!\vec{\hat n}_2]^T$, and $(\cdot)^T$ denote transpose of a vector or a matrix.
Since $\vec{\hat n}_1$ and $\vec{\hat n}_2$ are linearly independent, $V$ is invertible and thus we have $\vec{u} = V^{-T} \vec{d}$, where $(\cdot)^{-T}$ is inverse transpose.
From \cref{fig:theory_trianlge}, $d_1=\|\vec{u}\|\sin(\alpha_1)$ and $d_2=\|\vec{u}\|\sin(\alpha_2)$, where $\|\cdot\|$ denote the $\ell_2$-norm.
The motion direction is thus computed as
\begin{equation}
\vec{\hat{u}} = V^{-T}\begin{bmatrix}
    \sin(\alpha_1)\\\sin(\alpha_2)
\end{bmatrix}
\label{eq:motion vector direction compute}
\end{equation}
where $\hat u$ is the unit direction of $\vec u$.

We now relate the \eigvalues{} to $\sin(\alpha_1)$ and $\sin(\alpha_2)$.
We start with computing the structure tensor from \eqref{eq:structure_tensor_feature} at position $\vec{r}_{A'}$:
\begin{equation}
S_w(\vec{r}_{A'})=\left.w * (\vec{\nabla I} \vec{\nabla I}^T)\right|_{\vec{r}=\vec{r}_{A'}}=\int_{\vec{r}} w(\vec{r}_{A'} - \vec{r}) \,\vec{\nabla I}(\vec{r}) \vec{\nabla I}(\vec{r})^T d\vec{r},
\label{eq:structure matrix def}
\end{equation}
where $w(\cdot)$ is a window function, and vector $\vec{\nabla I}(\vec{r})$ is the gradient of density value image at $\vec{r}$ as described in \cref{sec:method subsec:feature extraction}.
As the triangle translates, the edge $AC$ drifts to $A'C'$, forming the trapezoidal region $AA'C'C$ where gradients align with $\vec{\hat n}_1$.
Similarly, in trapezoid $AA'B'B$, the gradients align with $\vec{\hat n}_2$.
Assuming a Gaussian window function, let $\Gamma$ be a circular region centered at $A'$ with radius $A'N$, outside of which the Gaussian response is negligible.
We can compute the structure tensor in \eqref{eq:structure matrix def} as
\begin{align}
S_w(\vec{r}_{A'})
  &= \sum_{k=1}^{2}\int_{\vec{r}\in R_k}
     w(\vec{r}_{A'}-\vec{r})\,\vec{\nabla I}(\vec{r})\,
     \vec{\nabla I}(\vec{r})^T d\vec{r} = \sum_{k=1}^{2} A_k\,\vec{\hat n}_k\vec{\hat n}_k^T \label{eq:open form structure matrix with normals}\\
   &= V\,\operatorname{diag}(A_1,A_2)\,V^T,
\label{eq:structure matrix with normals}
\end{align}
where $R_1$ and $R_2$ denote the corresponding sectors $A'PN$ and $A'PM$, and $A_1$, $A_2$ represent the relative strength of each gradient contribution.  
The equality between \eqref{eq:open form structure matrix with normals} and \eqref{eq:structure matrix with normals} is obtained by substituting $\vec{\hat n}_1$ and $\vec{\hat n}_2$ into matrix $V$.

\begin{algorithm}[t]
\caption{Computation of Motion Direction}
\label{alg:motion_direction}
\small
\begin{algorithmic}[1]
\Statex \textbf{Input:} Local geometry $V=\begin{bmatrix}\vec{\hat n}_1, \,\vec{\hat n}_2\end{bmatrix}$, Harris eigenvalues $\lambda_1$, $\lambda_2$
\Statex \textbf{Output:} Motion direction $\vec{\hat u}$
\State Compute $\theta = \pi - \arccos(\vec{\hat n}_1 \cdot \vec{\hat n}_2)$.
\State Compute $A_1$ and $A_2$ using \cref{thm:from eig to A}.
\State Compute $\vec{\hat u}$ using Eq.~\eqref{eq:motion direction from constants}.
\end{algorithmic}
\normalsize
\end{algorithm}

We assume that $A_1$ and $A_2$ are proportional to sector areas $R_1$ and $R_2$:
\begin{equation}
\frac{A_1}{A_2} = \frac{\mathrm{area}(R_1)}{\mathrm{area}(R_2)} = \frac{\alpha_1}{\alpha_2}.
\label{eq:constants to alphas}
\end{equation}
This assumption may be affected in practice by factors such as object texture, background variation, rasterization effect in computing the gradients, border effects, and  noise. \nergis{Very nice hypotheses! I think we should revisit this when we are testing other datasets, i.e. consider texture, background, noise, etc. when we test on `more complex' datasets. :)}
We will evaluate the effect of texture and noise in detail in \cref{sec:exp subsec:mlp-knn}.
Given \eqref{eq:constants to alphas} and \eqref{eq:motion vector direction compute}, the motion direction from constants $A_1$ and $A_2$ follows as
\begin{equation}
\vec{\hat u} = V^{-T}\!\left[\sin\!\tfrac{A_1\theta}{A_1+A_2},\;\sin\!\tfrac{A_2\theta}{A_1+A_2}\right]^T\!,
   \label{eq:motion direction from constants}
\end{equation}
where $\theta = \angle CAB$ is the angle at vertex $A$, computed as $\theta = \pi - \arccos(\vec{\hat n}_1 \cdot \vec{\hat n}_2)$.

\begin{theorem}
\label{thm:from eig to A}
Constants $A_1$ and $A_2$ can be computed from the eigenvalues of the structure tensor, $\lambda_1$ and $\lambda_2$, as the roots of the following quadratic equation:
\begin{equation}
    A^2 - (\lambda_1 + \lambda_2) A + \lambda_1 \lambda_2/\sin^2{\theta}=0.
    \label{eq:quadratic equation A from lmbda}
\end{equation}
 \noindent \textit{Proof:} See supplementary material.
\end{theorem}

Finally, \cref{alg:motion_direction} connects the combined Harris eigenvalues and the local geometric information to the motion direction, confirming the proposed link between eigenvalue features and motion cues.
 \nergis{Super nice derivation! I like it, and it should be in the paper as one of the main contributions. This way we can also put some variable names on our hypothesis, e.g. we hypothesize that local geometry (V)+eigenvalues (proportional to alphas) can give us motion direction (u). Very nice!}

\subsection{Toy data generation}
\label{sec:method subsec:toydatageneration}


\newcommand{\subfigureheight}{0.7\textwidth}

\begin{figure}[t]
    \centering
    \begin{subfigure}[b]{0.32\textwidth}
        \centering
        \includegraphics[width=\textwidth,
        height=\subfigureheight,
        keepaspectratio
    ]{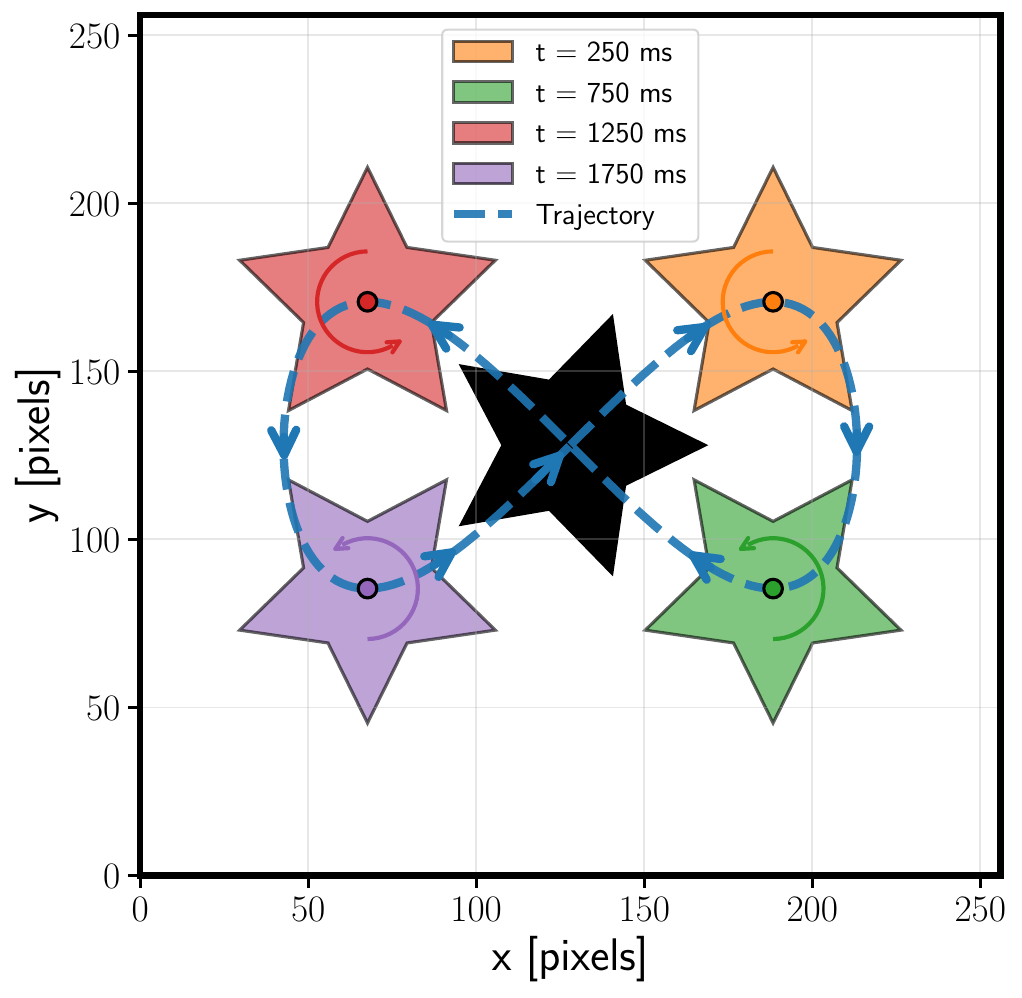}
        \caption{Star shape trajectory}
        \label{fig:toy_movement}
    \end{subfigure}
    \hfill
    \begin{subfigure}[b]{0.32\textwidth}
        \centering
        \includegraphics[        width=\textwidth,
        height=\subfigureheight,
        keepaspectratio
    ]{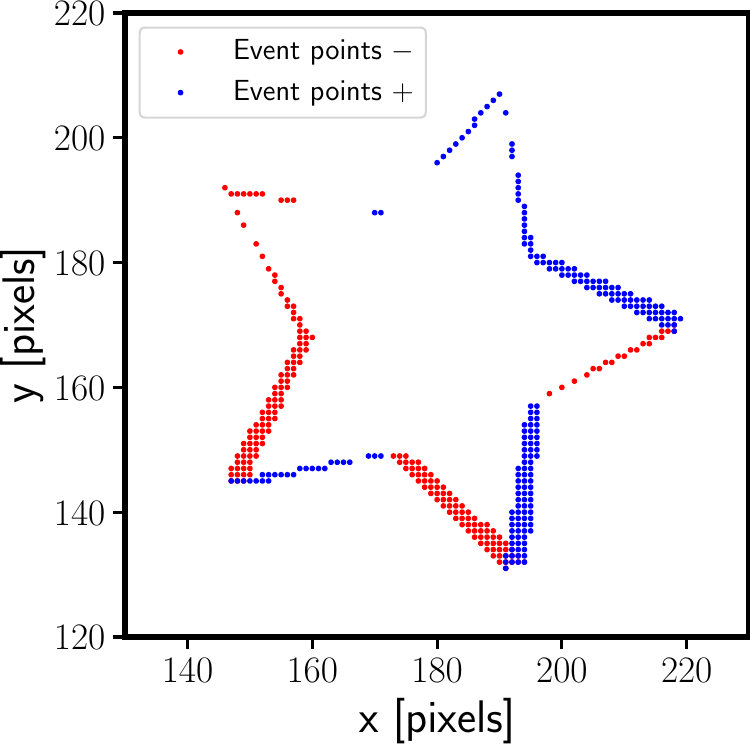}
        \caption{Events at $t{=}200$--$210$~ms}
        \label{fig:toy_events}
    \end{subfigure}
    \hfill
    \begin{subfigure}[b]{0.32\textwidth}
        \centering
        \includegraphics[width=\textwidth,
        height=\subfigureheight,
        keepaspectratio
    ]{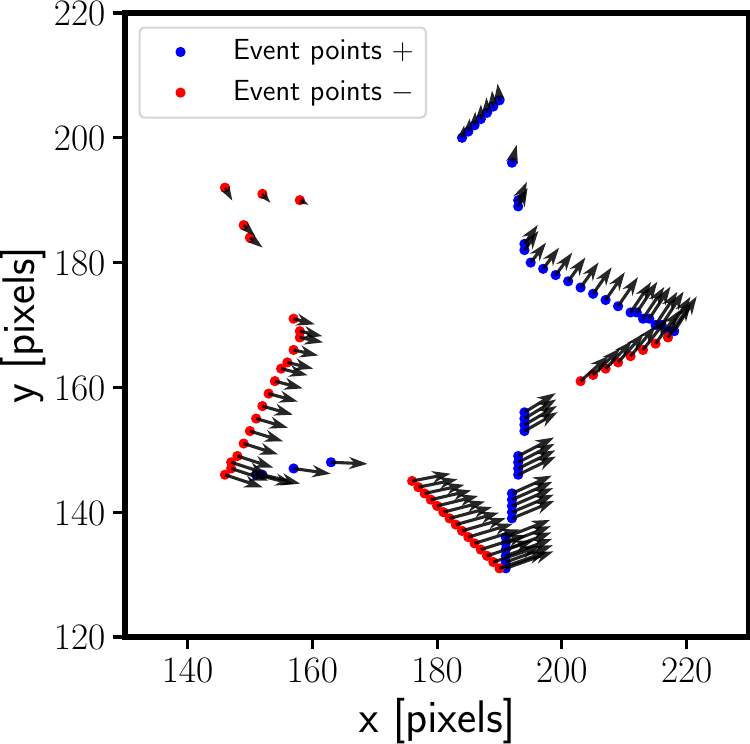}
        \caption{{{OF at {$t{=}200{-}201\,ms$}}}}
        \label{fig:toy_flow}
    \end{subfigure}   
    \begin{subfigure}[b]{0.32\textwidth}
        \centering
        \includegraphics[        width=\textwidth,
        height=\subfigureheight,
        keepaspectratio
    ]{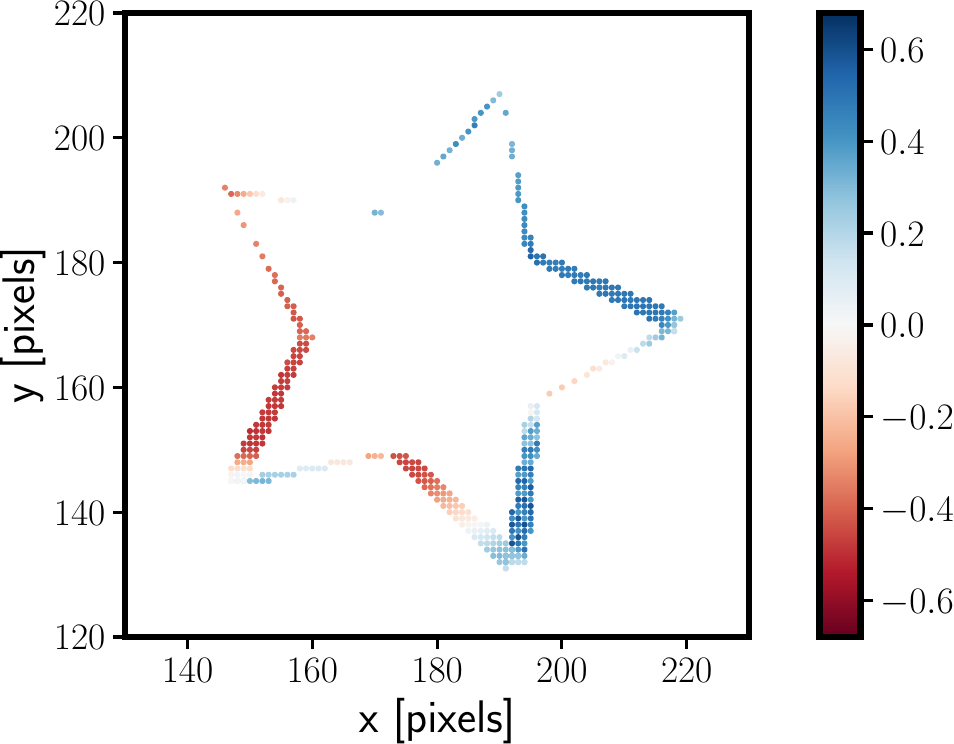}
        \caption{\Filtervalues{}}
        \label{fig:toy_density}
    \end{subfigure}
    \hfill
    \begin{subfigure}[b]{0.32\textwidth}
        \centering
        \includegraphics[        width=\textwidth,
        height=\subfigureheight,
        keepaspectratio
    ]{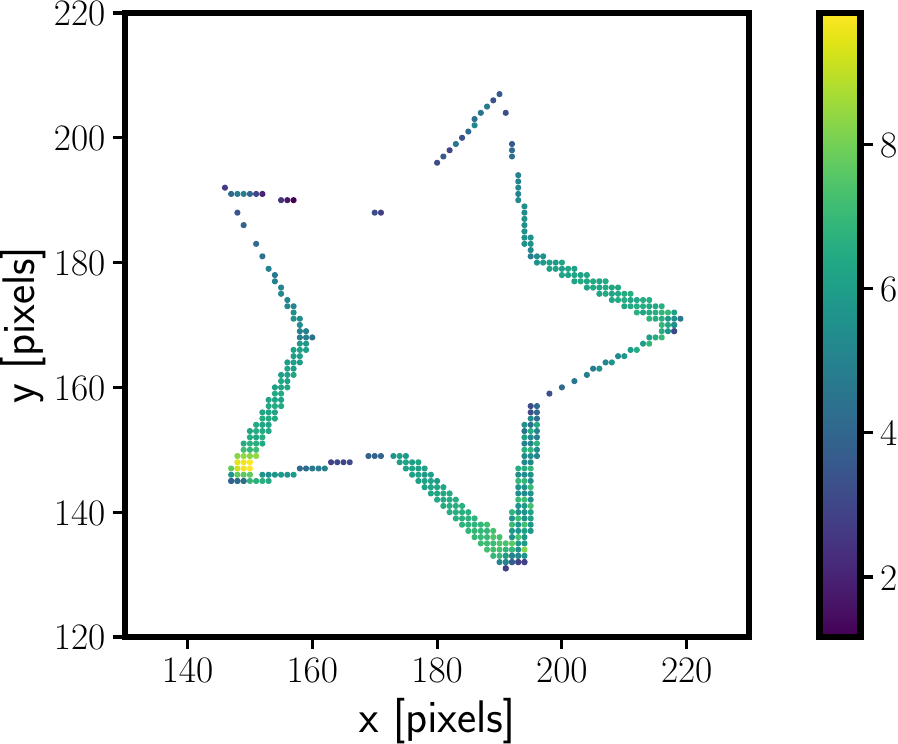}
        \caption{Eigenvalue $\lambda_1$}
        \label{fig:toy_eig1}
    \end{subfigure}
    \hfill
    \begin{subfigure}[b]{0.32\textwidth}
        \centering
        \includegraphics[        width=\textwidth,
        height=\subfigureheight,
        keepaspectratio
    ]{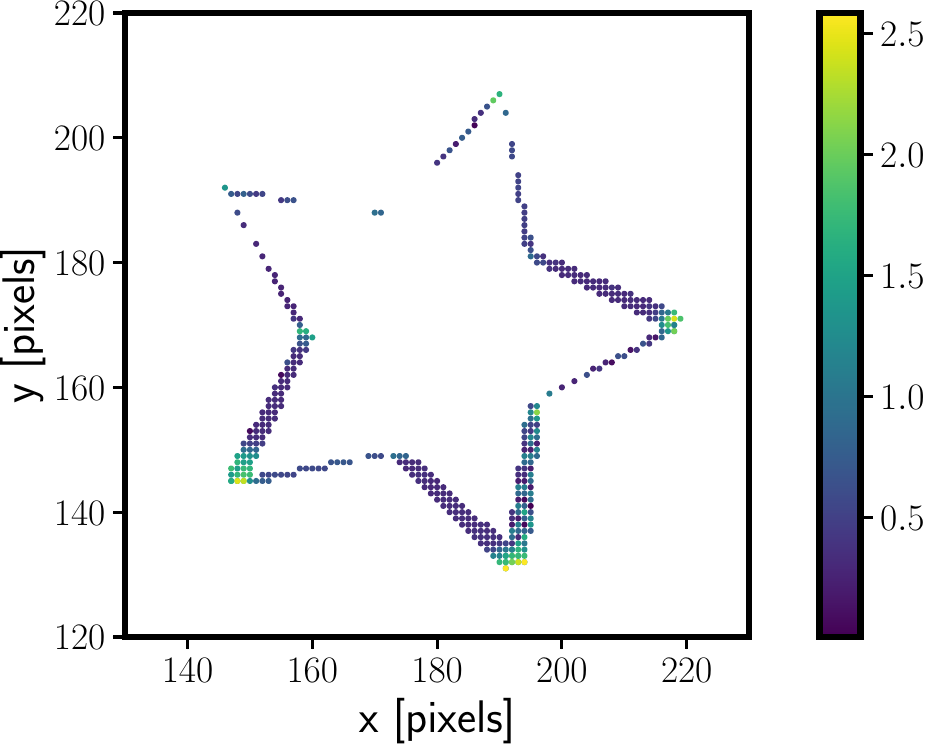}
        \caption{Eigenvalue $\lambda_2$}
        \label{fig:toy_eig2}
    \end{subfigure}
    \caption{%
        Toy dataset generation and feature extraction for a five-pointed star shape undergoing simultaneous translation and rotation.
        \textbf{(a)} The star follows a figure-eight (lemniscate) trajectory while rotating, with colored instances showing positions at different time steps and curved arrows indicating local rotation.
        \textbf{(b)} Events generated during a $10$~ms window, with blue ($+$) and red ($-$) polarities indicating brightness changes at object edges.
        \textbf{(c)} Ground truth optical flow (OF) vectors overlaid on events, showing the combined translational and rotational motion.
        \textbf{(d)--(f)} Per-event features computed from the event stream: \filtervalues{} capturing spatiotemporal event density, and \eigvalues{} $\lambda_1$ and $\lambda_2$.
    }
    \label{fig:toy_dataset_generation}
\end{figure}

To evaluate our proposed features in a controlled setting, we design a synthetic toy dataset where ground truth optical flow is available for each event.
We simulate a five-pointed star shape undergoing simultaneous translation and rotation within a $256 \times 256$ pixel image space.
The shape follows a figure-eight (lemniscate of Gerono) trajectory defined by $x(t) = A\sin(t)$ and $y(t) = B\sin(2t)$, while simultaneously rotating with a constant angular velocity, completing two full rotations over the sequence duration. \nergis{Can we cite the paper where we got this idea from here? If others use the same shape/motion for the synthetic star dataset, it's not nice not to cite them.}\hesam{I do not remember we get the idea from somewhere?}

For each event, we compute the ground truth optical flow vector from the known instantaneous velocity of the shape, combining both the translational component from the trajectory and the rotational component around the shape's center.
We reserve the last 20\% of the sequence for testing and use the first 80\% for training and validation.
\Cref{fig:toy_dataset_generation} illustrates the dataset generation process and the extracted features.
From the generated events, we compute the per-event features described in \cref{sec:method subsec:feature extraction}: the \filtervalues{} $I$ and the structure tensor \eigvalues{} $(\lambda_1, \lambda_2)$.
As shown in \cref{fig:toy_density}--\ref{fig:toy_eig2}, 
the \filtervalues{} captures the density of recent events, 
first eigenvalue $\lambda_1$ is generally larger at edge regions, while the second eigenvalue $\lambda_2$ is higher at corner regions.

\subsection{Models}
\label{sec:method subsec:model explanation}


\noindent\textbf{\Toymodel}
We design a toy model for supervised regression on the toy dataset to test whether \eigvalues{} and density features contain information to predict motion.
For each event, we compute the features as described in \cref{sec:method subsec:feature extraction} and use them as input to the model.
Each event $e_i$ has a ground truth flow vector $\vec{v}_i \in \mathbb{R}^{2}$.
To balance spatial and temporal distances, we scale the time in milliseconds. 
We define a spatiotemporal local neighborhood set $\mathcal{N}(i)$ around event $e_i$ by a $K$ nearest neighbor set of events.
The goal of the model is to predict $\hat{\vec v}_i$ from the features of event $e_i$ and $\mathcal{N}(i)$.
Each event $e_i$ is assigned a feature vector $\vec{z}_i\in\mathbb{R}^{F}$.
For all neighbors $j\in\mathcal{N}(i)$, we use relative spatial coordinates $\vec{p}_j{=}[\Delta x_j, \Delta y_j]^T = [x_j - x_i, y_j - y_i]^T$ instead of absolute positions.
This design choice prevents the model from overfitting by memorizing flow vectors based on absolute pixel locations, and provides the local spatiotemporal geometry of events in the neighborhood of the center event.

\begin{figure}[t]
    \centering
    \includegraphics[width=0.75\linewidth]{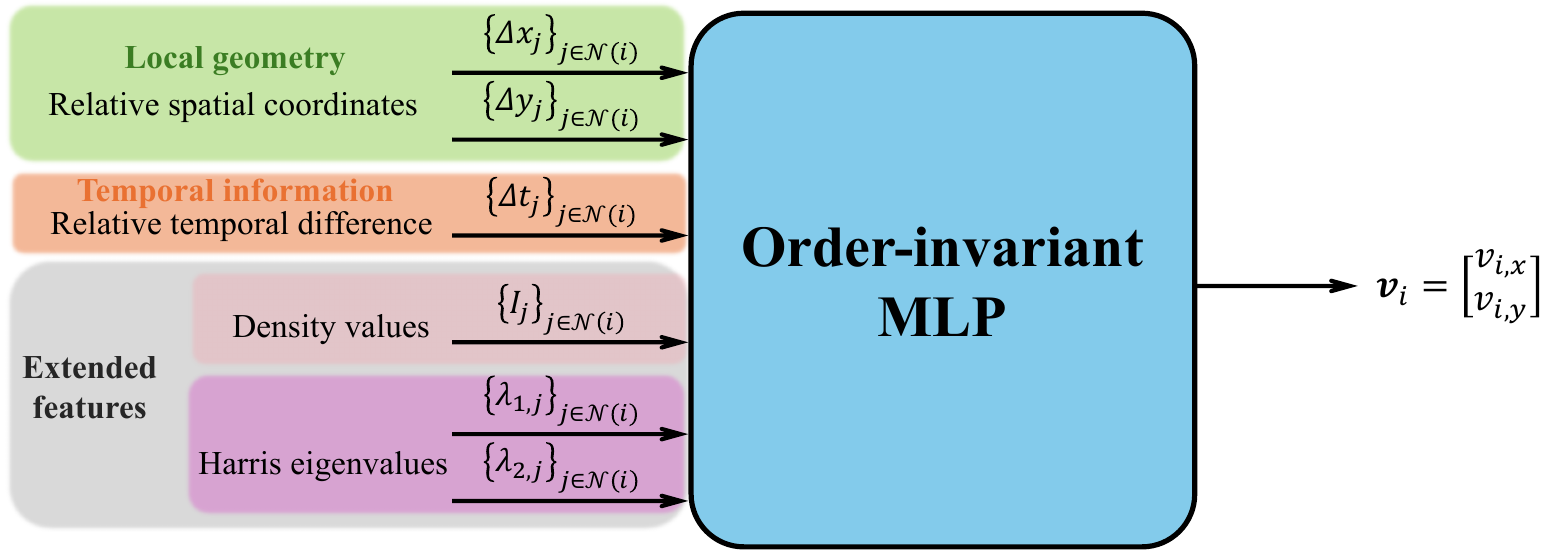}
    \caption{Input features of the \toymodel{}: The model can receive per-event information for local geometry (relative spatial coordinates), temporal information (relative temporal difference), and extended features (density values and Harris eigenvalues), and outputs the optical flow vector.}
    \label{fig:feature_types}
\end{figure}

We consider the following feature configurations:
{\Originalcase{}} ${\vec z_j = \vec{p}_j}$;
{\Eigvalues{}} ${\vec z_j = [\vec{p}_j,\lambda_{1,j},\lambda_{2,j}]}$;
{\Filtervalues{}} ${\vec z_j = [\vec{p}_j,I_j]}$; and
{\bothshortvalues{}} ${\vec z_j = [\vec{p}_j,I_j,\lambda_{1,j},\lambda_{2,j}]}$,
where $(\lambda_{1,j},\lambda_{2,j})$ are the Harris eigenvalues and $I_j$ is the density value from \eqref{eq:filter_value}.
We also consider time-extended variants where $\Delta t_j$ is included in $\vec{p}_j$, \ie, $\vec{p}_j=[\Delta x_j, \Delta y_j, \Delta t_j]^T$.
\Cref{fig:feature_types} illustrates the different input features to the \toymodel{}.

The neighborhood $\mathcal{N}(i)$ is a set; its ordering should not affect the prediction.
We therefore use a permutation-invariant architecture inspired by DeepSets~\cite{zaheer2017deepsets}:
\begin{equation}
\hat{\vec v}_i = \mathrm{MLP}_2\Big(\sum_{j\in\mathcal{N}(i)\setminus\{i\}}\mathrm{MLP}_1\left(\vec z_i\|\vec z_j\right)\Big),
\label{eq:toymodel_architecture}
\end{equation}
where $\|$ denotes concatenation.
A shared $\mathrm{MLP}_1$ embeds each pair of central and neighbor features, the sum aggregation ensures permutation invariance, and $\mathrm{MLP}_2$ produces the final flow prediction.
Each of the MLPs has two hidden layers with the same number of units and ReLU activations.
We train the model with mean-squared error over all events. 

\noindent\textbf{Iterative Deblurring Network (IDNet)}
To evaluate our proposed features on a state-of-the-art \emph{supervised} optical flow estimation method, we adopt the Iterative Deblurring Network (IDNet)~\cite{wu_lightweight_2024}.
IDNet achieves competitive performance while remaining lightweight by avoiding the construction of costly correlation volumes used in methods like ERAFT~\cite{gehrig_e-raft_2021}.
IDNet uses a voxel grid representation of events with a convolutional encoder at the input level.
To test our hypothesis that \eigvalues{} and \filtervalues{} aid optical flow estimation, we voxelize the proposed per-event features into the same temporal bins and concatenate them with the event voxels along the channel dimension.
See supplementary material for architecture and training details.


\section{Experiments}
\nergis{Potential reviewer questions about more experiments: 1. More datasets (e.g. MVSEC), 2. More models (other than just IDNet), 3. More baseline representations with IDNet/tinyIDNet (e.g. compare against time surface, or tencode, etc. in an extra channel vs. eig.+density in extra channel. For the current DSEC experiments (Table 4/5), they might also ask for: more reruns with different seeds, more different resolutions (e.g. 1/2 res?), more sub-datasets (what if you use only 1 sequence or 2 sequences for training instead of 4?), more hyperparameter searches/ablations on the IDNet (e.g. with different number of iterations in the recurrent network, etc.)... We don't have to do all of these, but let's keep it in mind.}
\label{sec:exp}

\subsection{Toy dataset with an \toymodel}
\label{sec:exp subsec:mlp-knn}

\remove{In this experiment, we construct a synthetic dataset, as explained in \cref{sec:method subsec:toydatageneration}, using the star-shaped foreground object whose motion follows a Lemniscate of Gerono trajectory.
From this known motion pattern, we generate event streams and compute the corresponding per-event motion vector flow fields, which serve as ground-truth supervision.}
\movesupp{We train for 500 epochs using Adam with an MSE objective, learning rate $10^{-3}$, batch size 1024, and a OneCycle learning-rate schedule, and report results over 10 random seeds.}
\nergis{My suggestion is to make a figure with the model setups (e.g. the different kinds of input we give the MLP etc.) and clearly mark in the figure which one is "augmented", and also explain it in text what we mean by augmented.}


\begin{figure*}[t]
  \centering
  \begin{subfigure}[t]{0.32\textwidth}
    \centering
    \includegraphics[width=\linewidth]{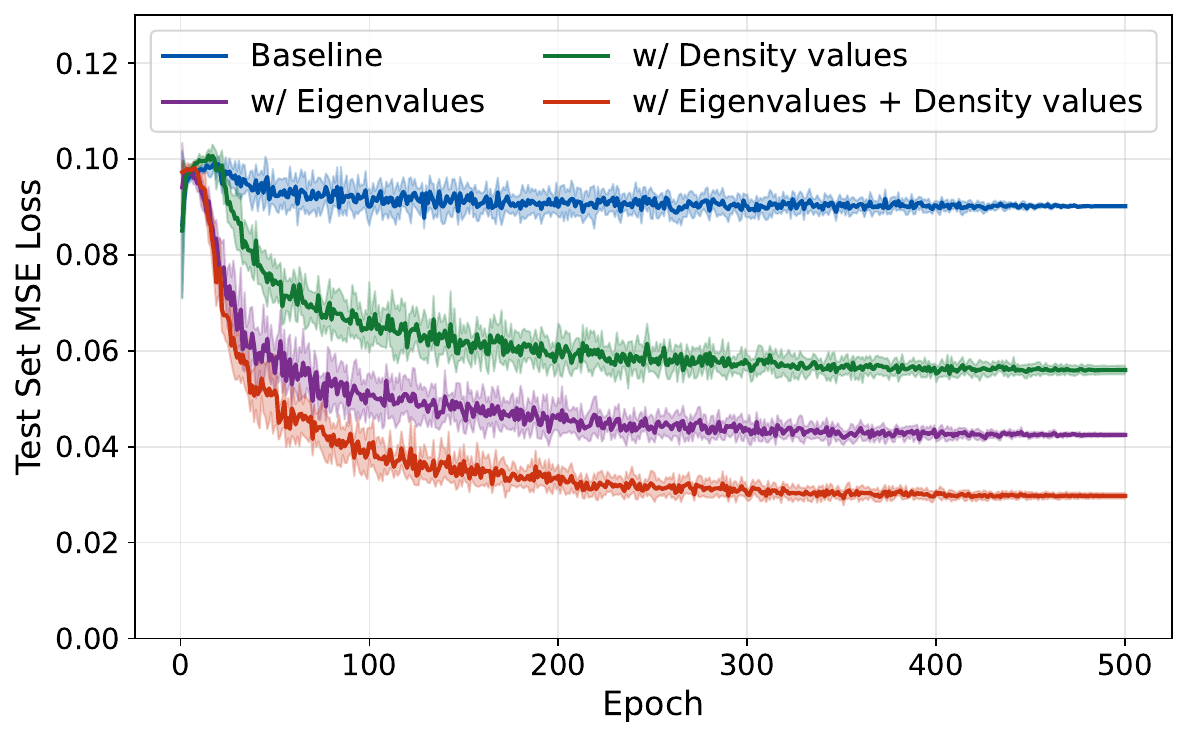}
    \caption{hidden\_dim=64, k=5}
  \end{subfigure}\hfill
  \begin{subfigure}[t]{0.32\textwidth}
    \centering
    \includegraphics[width=\linewidth]{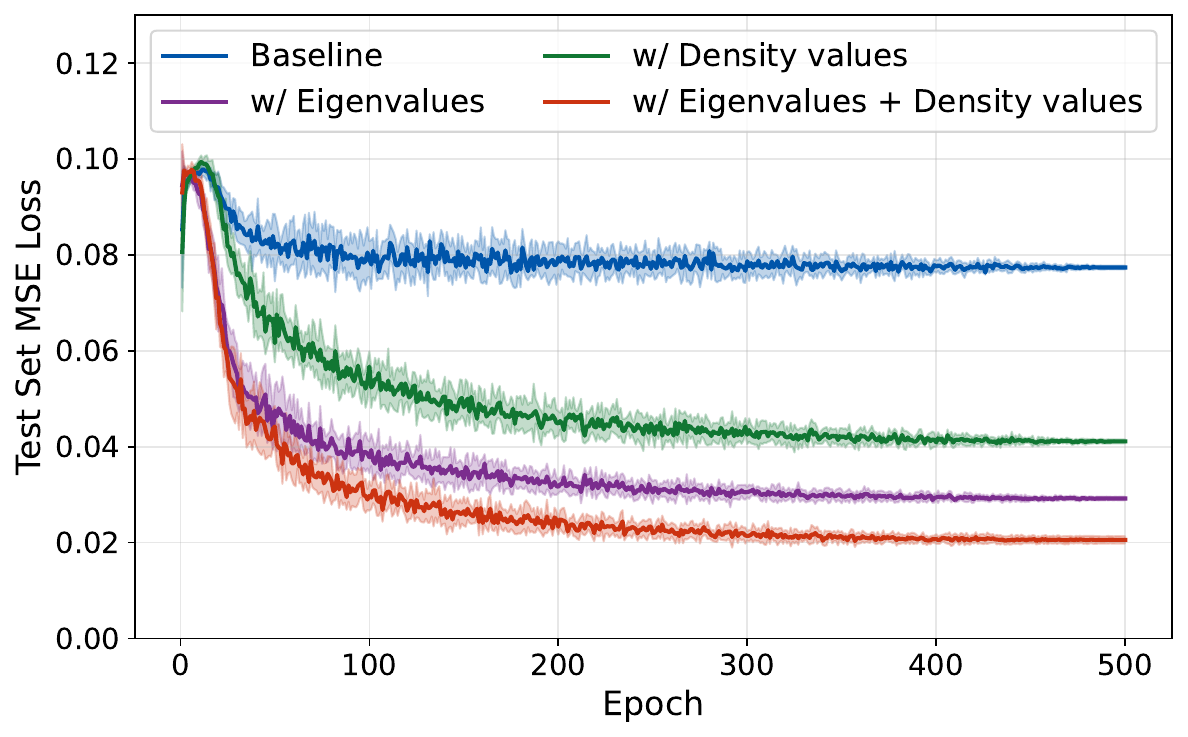}
    \caption{hidden\_dim=64, k=10}
  \end{subfigure}\hfill
  \begin{subfigure}[t]{0.32\textwidth}
    \centering
    \includegraphics[width=\linewidth]{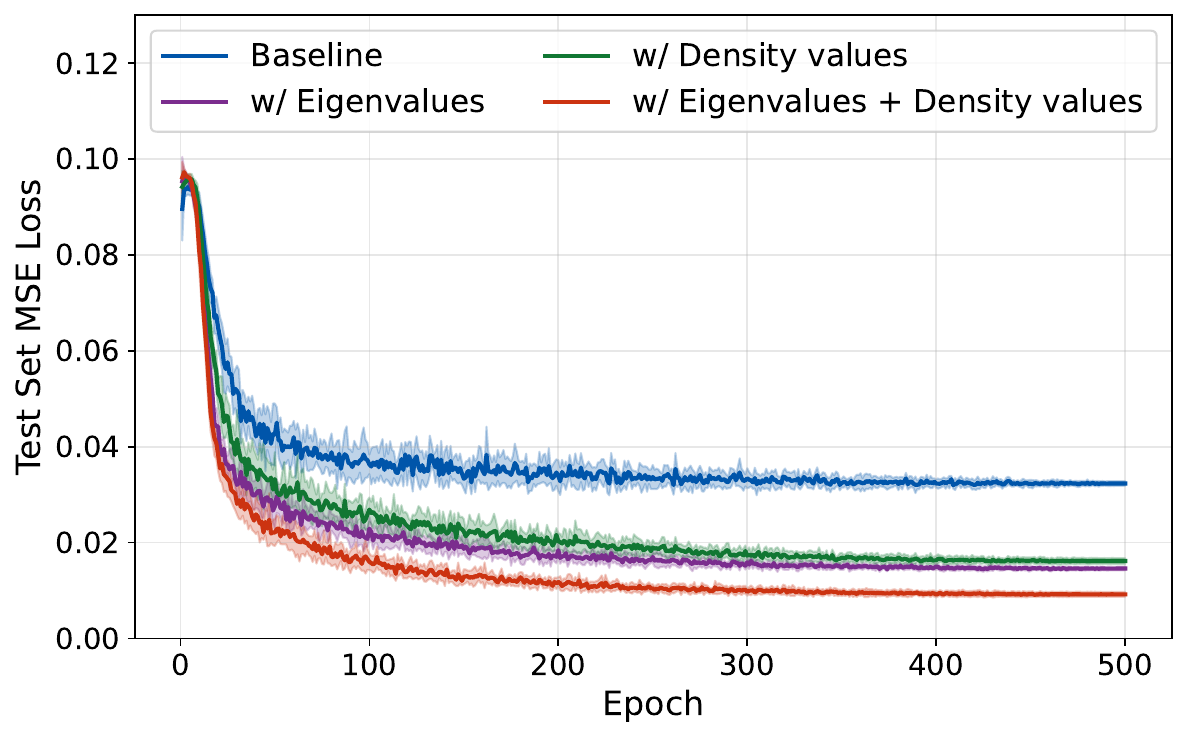}
    \caption{hidden\_dim=64, k=50}
  \end{subfigure}
  \begin{subfigure}[t]{0.32\textwidth}
    \centering
    \includegraphics[width=\linewidth]{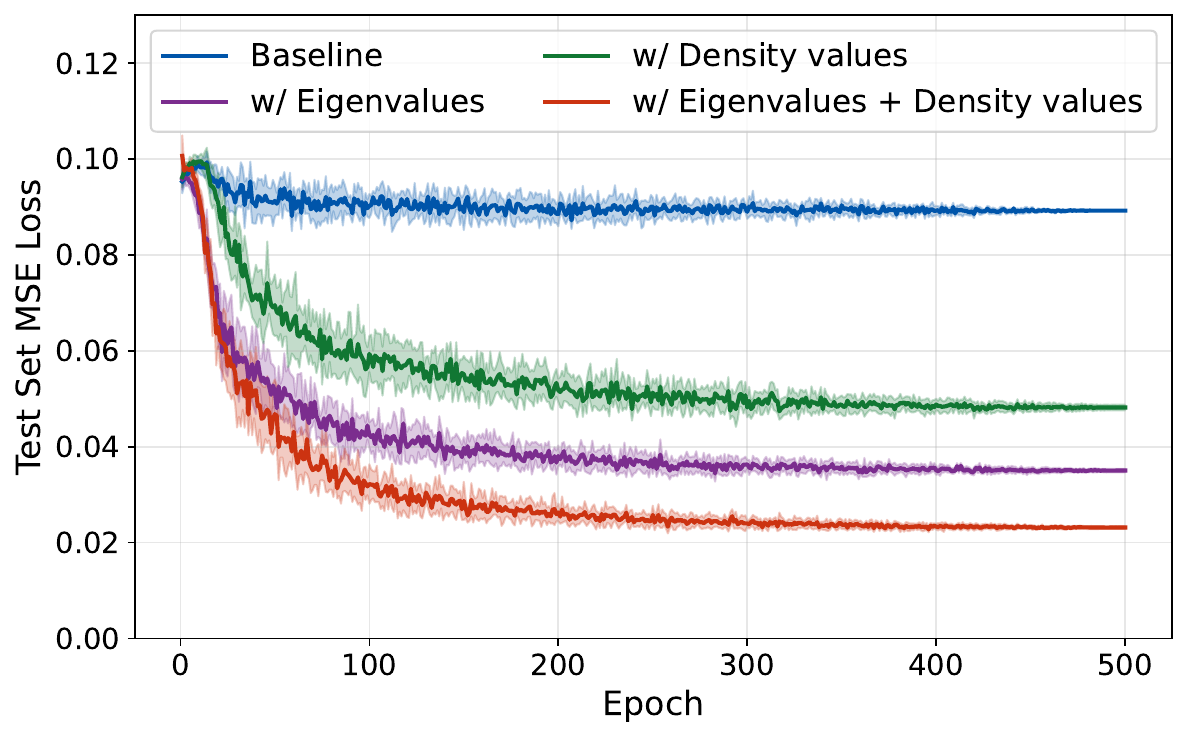}
    \caption{hidden\_dim=128, k=5}
  \end{subfigure}\hfill
  \begin{subfigure}[t]{0.32\textwidth}
    \centering
    \includegraphics[width=\linewidth]{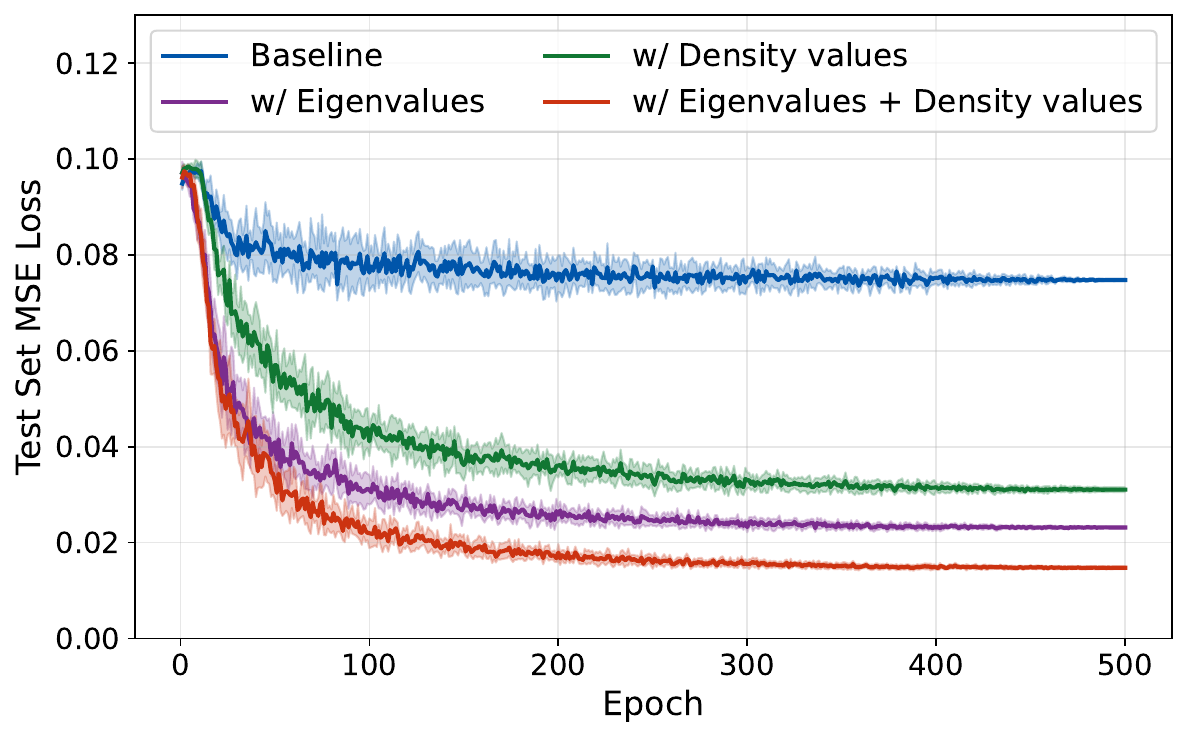}
    \caption{hidden\_dim=128, k=10}
  \end{subfigure}\hfill
  \begin{subfigure}[t]{0.32\textwidth}
    \centering
    \includegraphics[width=\linewidth]{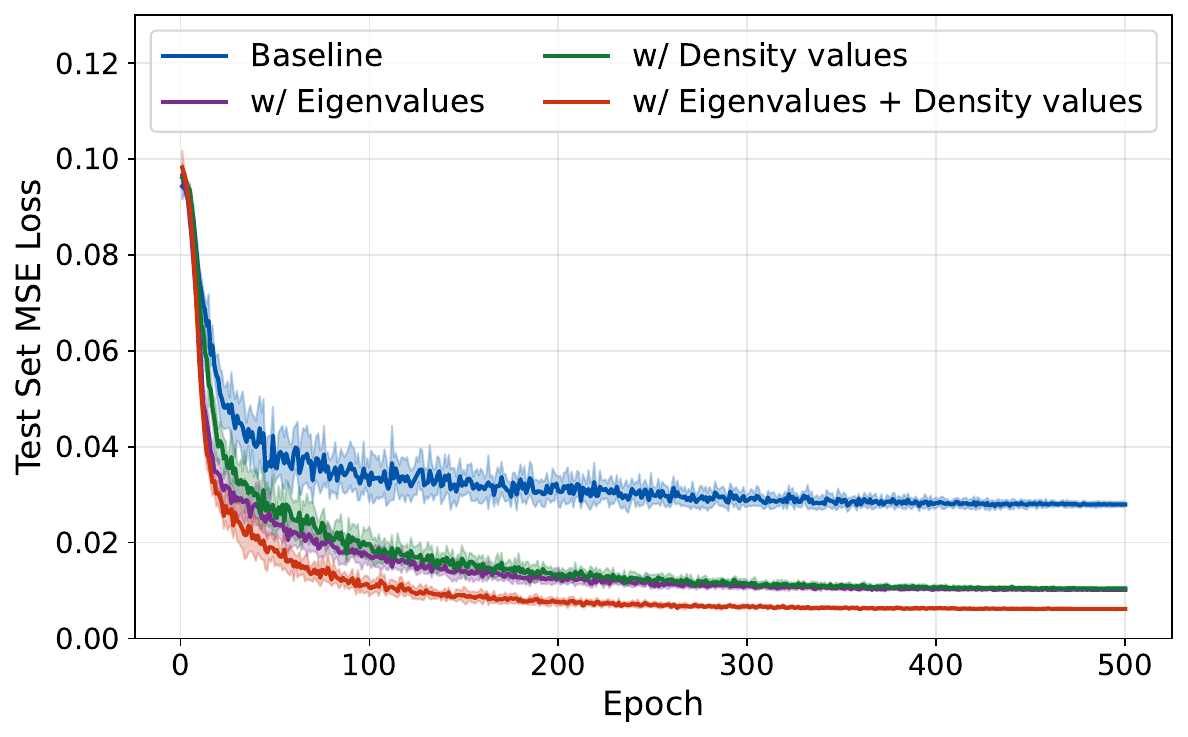}
    \caption{hidden\_dim=128, k=50}
  \end{subfigure}
  \caption{Test set MSE loss over training epochs comparing \originalcase{} to extended inputs (\Eigvalues{}, \Filtervalues{}, and both) across different hidden dimensions and $k$-nearest neighbors. Shaded areas indicate $\pm 1$ standard deviation across 10 runs.}
  \label{fig:toy:hidden_and_k}
\end{figure*}

\noindent\textbf{Do the additional \eigvalues{} and \filtervalues{} features provide motion cues?}
First, we test the hypothesis that \eigvalues{} and density features, when combined with local geometry, provide motion cues on the toy dataset.
We evaluate the four input variants, 
\originalcase{}, \eigvalues{}, \filtervalues{}, and \bothshortvalues{}, across two MLP hidden dimensions (64 and 128) and three neighborhood sizes $K \in \{5, 10, 50\}$.
\Cref{fig:toy:hidden_and_k} shows MSE loss over training epochs, and additional metrics are provided in Supplementary.

Overall, adding the extended features consistently reduces the test MSE loss compared to using relative positions alone, with best performance observed with both eigenvalues and density features.
Increasing $K$ or hidden dimensions benefit all feature settings, with larger gains observed for density features.
This shows that density features along with larger neighborhoods, i.e. richer local geometry, can yield better results in capturing motion patterns.
Across all tested configurations, combining \eigvalues{} and \filtervalues{} features produces the lowest test MSE loss.


\begin{figure}[t]
  \centering
  \begin{subfigure}[t]{0.4\linewidth}
    \centering
    \includegraphics[width=\linewidth]{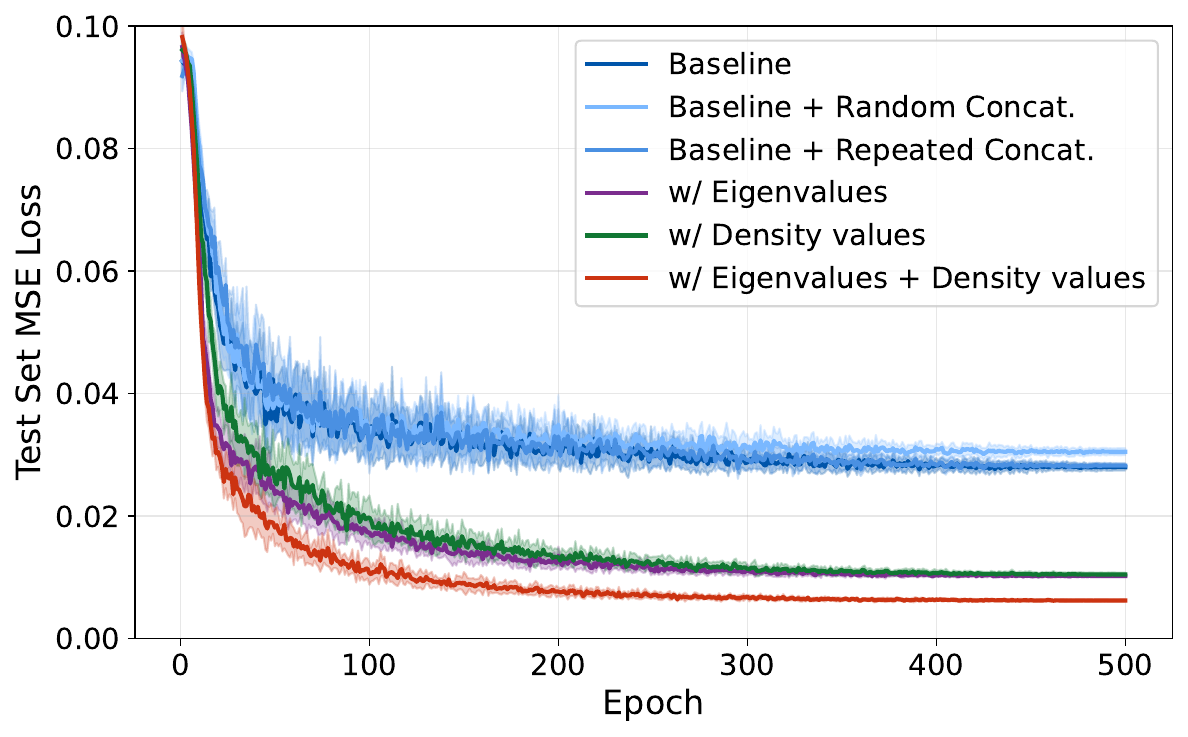}
    \caption{Matching input dimensionality}
    \label{fig:toy:repeated_random_extending}
  \end{subfigure}
\begin{subfigure}[t]{0.4\linewidth}
  \centering
  \includegraphics[width=\linewidth]{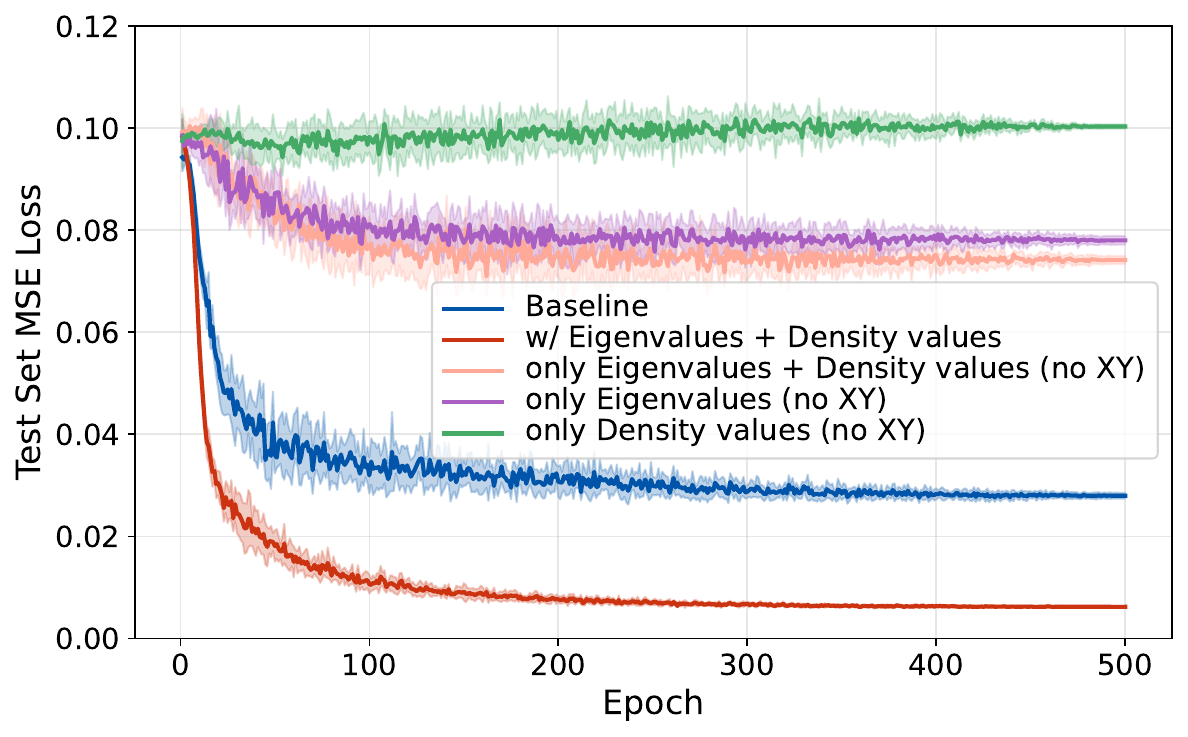}
  \caption{Excluding relative position}
  \label{fig:toy:exclude_relative_position}
  \end{subfigure}
  \caption{Test set MSE loss on the toy dataset over 10 runs when \textbf{(a)} matching \originalcase{} input dimensionality via random or repeated position features. \bothshortvalues{} still outperforms these controls, confirming gains stem from informative features, not increased dimensionality. \textbf{(b)} Excluding relative position ($\Delta x,\Delta y$) causes models using only \eigvalues{} or \filtervalues{} features to fail.
  }
  \label{fig:toy:controls}
\end{figure}

\noindent\textbf{Are the gains due to informative features or merely increased model capacity?}
A natural concern is that extended features increase the input dimensionality (and thus the number of parameters), potentially improving performance even with uninformative features. \nergis{This is a good example of a motivation sentence, it would be great to more or less start each section with a sentence like this, i.e. what is the `concern'? what do we want to `control' in this experiment? what is the hypothesis (e.g. since relative time is used in the eigenvalue computation, maybe adding time information as input would cancel out the effect of the eig.+density augmentation?, etc.) This is what I mean by `why'? :)} To control for this, we match the \bothshortvalues{} input dimension by either concatenating random features or repeating the relative-position features. \Cref{fig:toy:repeated_random_extending} shows that neither control reduces the test MSE, whereas extending with \eigvalues{}, \filtervalues{}, or both yields clear gains, indicating that the improvements stem from informative features rather than increased capacity.


\remove{We hypothesize that the \eigvalues{} and \filtervalues{} in combination with local geometry (relative positional) can provide extra information about motion flows.}
\noindent\textbf{Are extended features sufficient on their own without local geometry?}
As it was discussed in \cref{sec:method subsec:triangle theory}, the \eigvalues{} features along with local geometry can be used to recover the motion direction.
To test the necessity of local geometry, we trained models that exclude relative spatial coordinates ($\Delta x_j,\Delta y_j$).
\Cref{fig:toy:exclude_relative_position} shows that these models fail to learn optical flow.
This emphasizes our hypothesis that \eigvalues{} and \filtervalues{} features are motion cues when complemented with local geometry.
Moreover, this experiment shows that the model does not overfit on the extended features or memorize the motion flows derived from them.

\begin{figure*}[t]
  \centering
  \begin{subfigure}[t]{0.33\textwidth}
    \centering
    \includegraphics[width=\linewidth]{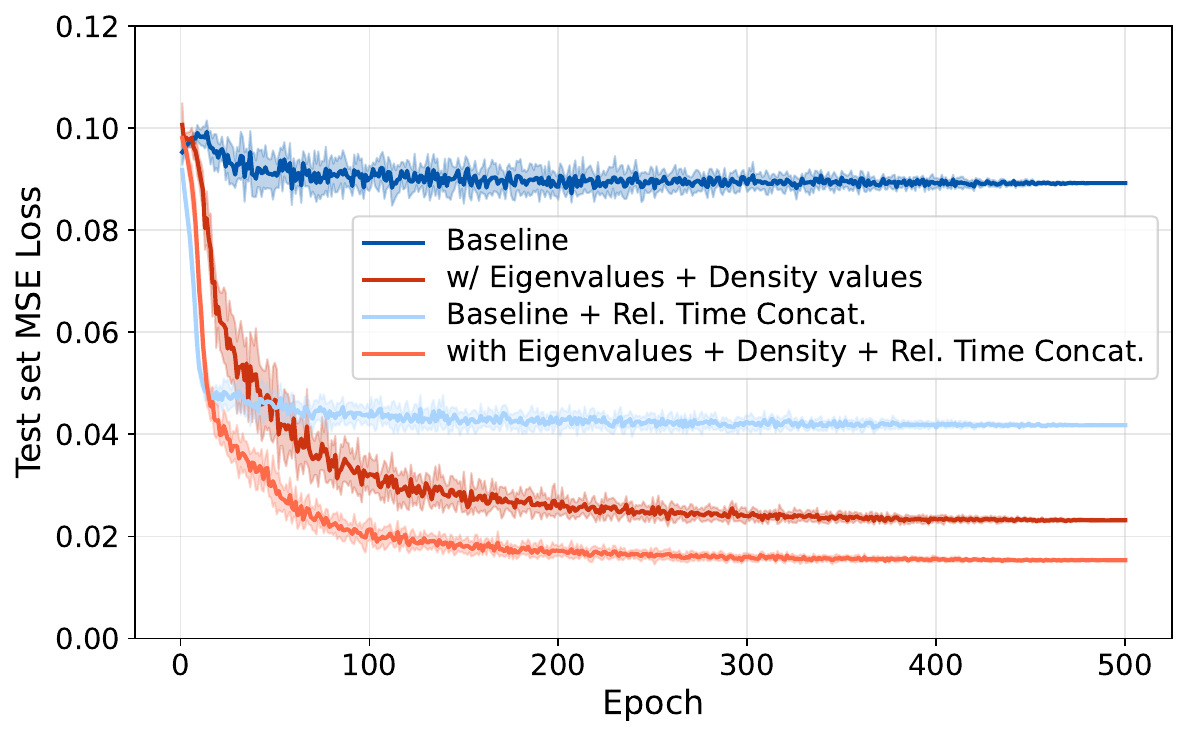}
    \caption{k=5}
    \label{fig:toy:timeaug_k5}
  \end{subfigure}\hfill
  \begin{subfigure}[t]{0.33\textwidth}
    \centering
    \includegraphics[width=\linewidth]{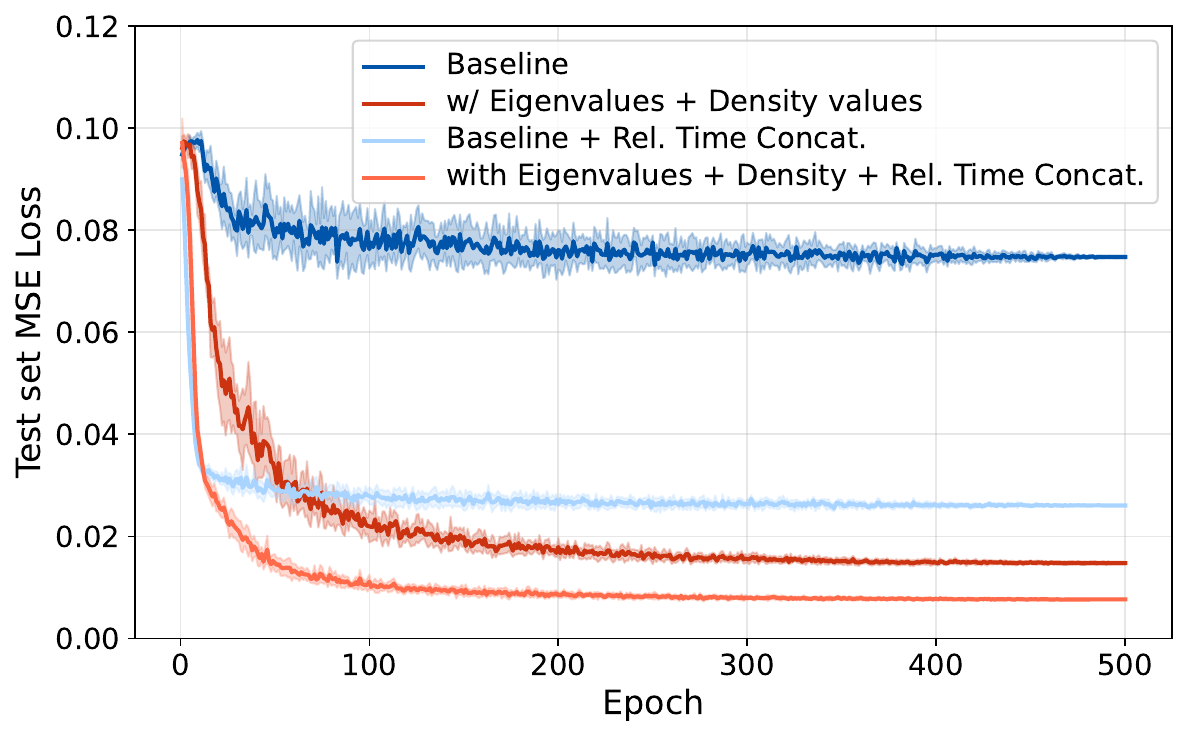}
    \caption{k=10}
    \label{fig:toy:timeaug_k10}
  \end{subfigure}\hfill
  \begin{subfigure}[t]{0.33\textwidth}
    \centering
    \includegraphics[width=\linewidth]{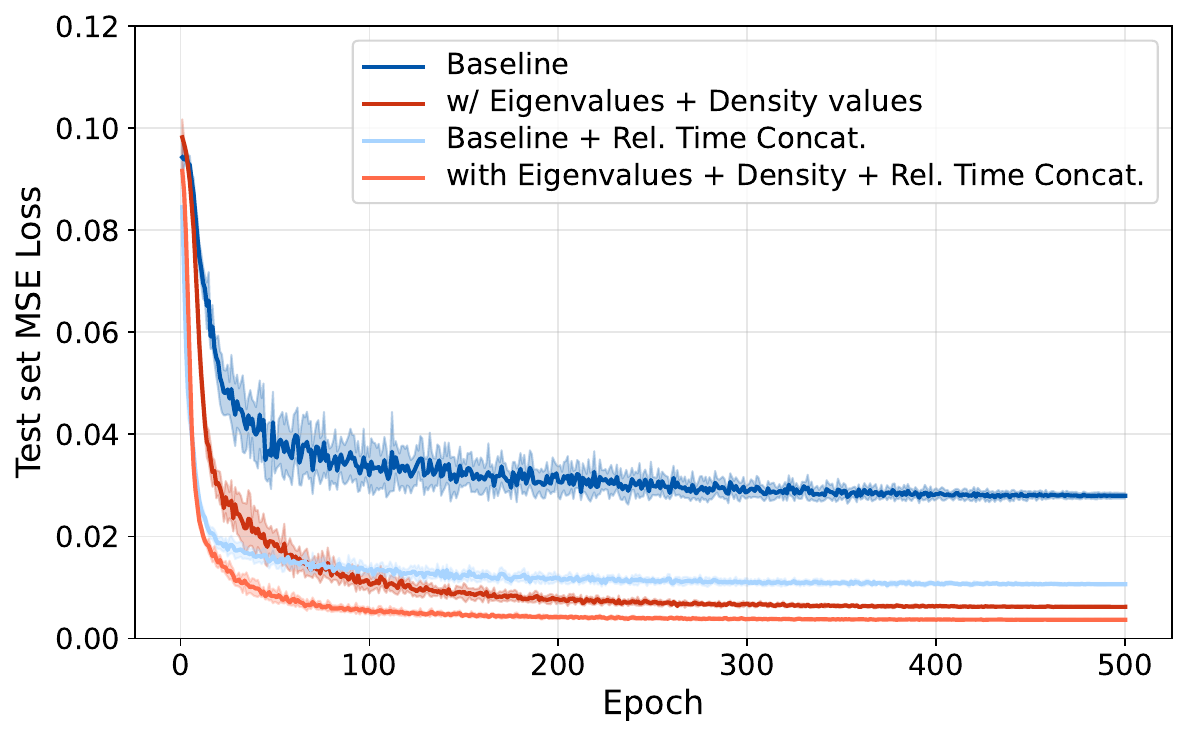}
    \caption{k=50}
    \label{fig:toy:timeaug_k50}
  \end{subfigure}
  \caption{Relative time as an additional input feature for different $K$. Combined \eigvalues{} and \filtervalues{} features provide more informative features for optical flow than relative time.}
  \label{fig:toy:timeaug_k}
\end{figure*}


\noindent\textbf{Do \eigvalues{} and \filtervalues{} features add information beyond relative time?}
Since relative temporal information is used in computing the extended features as in \cref{sec:method subsec:feature extraction}, one may argue that simply appending relative time to the input could capture the same motion cues as \eigvalues{} and \filtervalues{}.
To control for this, we compare \bothfullvalues{} against a time-extended baseline. 
\Cref{fig:toy:timeaug_k} shows results across different $K$.
Although time improves the baseline, \bothfullvalues{} achieve lower test MSE across all $K$, and adding time on top of \bothfullvalues{} yields further gains.
As $K$ increases, the performance gap narrows, %
yet the time-extended baseline still does not match the extended feature sets.


\noindent\textbf{Feature Extending Under Textured Conditions}
As we mentioned in \cref{sec:method subsec:triangle theory}, the relation between the \eigvalues{} and the direction of the movement can be affected by factors like object textures. To study this effect, we increase the complexity of the toy dataset by adding different texture patterns to the moving foreground object.
Textures are randomly selected from the Describable Textures Dataset (DTD)~\cite{cimpoi14describing}.
\Cref{tab:toy:texture_results} reports the test MSE loss for each texture, averaged over 10 runs with standard deviations. 
 The results show that incorporating \eigvalues{} and \filtervalues{} features consistently improves motion estimation in textured scenarios. 
MSE loss curves, additional metrics, and visualizations of the toy dataset with textures are provided in Supplementary.

\begin{table}[t]
\centering
\caption{Mean squared error (MSE) loss for different texture settings using the \toymodel{} with baseline and combined extended features. For each texture, the mean and standard deviation over 10 independent runs are reported.}
\label{tab:toy:texture_results}
\begin{adjustbox}{width=\textwidth}%
\begin{tabular}{lccccccc}
\hline
Feature Type & stratified-0106 & flecked-0103 & waffled-0063 & crosshatched-0066 & woven-0053 & spiralled-0137 & Overall (Mean $\pm$ Std) \\
\hline
Baseline & $0.0886 \pm 0.0002$ & $0.0799 \pm 0.0003$ & $0.0779 \pm 0.0002$ & $0.0762 \pm 0.0002$ & $0.0895 \pm 0.0003$ & $0.0644 \pm 0.0005$ & $0.0794 \pm 0.0084$ \\
w/ Eig. + Density & $0.0711 \pm 0.0005$ & $0.0668 \pm 0.0010$ & $0.0583 \pm 0.0009$ & $0.0540 \pm 0.0008$ & $0.0467 \pm 0.0004$ & $0.0411 \pm 0.0008$ & $0.0565 \pm 0.0105$ \\
Improvement(\%) & 19.74\% & 16.41\% & 25.10\% & 29.16\% & 47.75\% & 36.19\% & 28.84\% \\
\hline
\end{tabular}

\end{adjustbox}
\end{table}

\noindent\textbf{Feature Extending Under Shot Noise Events}
Beyond texture, noise is another practical factor that can degrade the proposed features. We evaluate robustness by injecting shot noise events at varying frequencies.
The noise frequencies tested are 0, 5, 20, and 50~Hz, where 0~Hz corresponds to no added noise. 
 \Cref{tab:toy:shot noise} reports the test MSE loss for each noise frequency, averaged over 10 runs.
The results show that the model extended with \bothfullvalues{} maintains higher accuracy and exhibits stronger resilience to increasing shot noise levels. 
MSE loss curves, comparisons in different metrics, and visualizations under different noise levels are provided in Supplementary.

\begin{table}[t]
\centering
\caption{Test MSE loss under added shot noise with different frequencies (0~Hz indicates no added noise). Results are obtained using the \toymodel{}. For each setting, the mean and standard deviation over 10 independent runs are shown.}
\label{tab:toy:shot noise}
\begin{adjustbox}{width=0.9\textwidth}%
\begin{tabular}{lccccc}
\hline
Feature Type & 0 Hz & 5 Hz & 20 Hz & 50 Hz & Overall (Mean $\pm$ Std) \\
\hline
Baseline & $0.0282 \pm 0.0004$ & $0.0361 \pm 0.0008$ & $0.0480 \pm 0.0008$ & $0.0606 \pm 0.0004$ & $0.0434 \pm 0.0124$ \\
w/ Eig. + Density & $0.0062 \pm 0.0001$ & $0.0104 \pm 0.0002$ & $0.0130 \pm 0.0006$ & $0.0196 \pm 0.0005$ & $0.0122 \pm 0.0049$ \\
Improvement(\%) & 78.08\% & 71.20\% & 72.83\% & 67.67\% & 71.99\% \\
\hline
\end{tabular}
\end{adjustbox}
\end{table}

\noindent\textbf{Toy dataset with IDNet}
So far, we employed an \toymodel{}, to measure the per-event ``motion information'' the \eigvalues{} and \filtervalues{}s contain. 
In this section, we evaluate the effectiveness of the proposed features in a more realistic scenario, using
a state-of-the-art optical flow architecture. \nergis{we don't have to say state of the art per se, but we're missing some justification here about why IDNet, e.g. it's a modern, popular, efficient, successful architecture for event-based OF or sth...?}
Specifically, we use a reduced-capacity configuration of IDNet~\cite{wu_lightweight_2024}, which we refer to as \emph{Tiny IDNet}.
We reduce the hidden-layer size and the RNN input dimensionality to match the low complexity of our toy dataset.
Unlike the \toymodel{}, which predicts per-event flow vectors, Tiny IDNet predicts frame-to-frame optical flow (between consecutive ground-truth frames) over short temporal windows.
We segment the event stream into temporal chunks of 20~ms and estimate one optical-flow field per chunk from its events.


\begin{figure}[t]
  \centering
  \begin{subfigure}[t]{0.45\textwidth}
    \centering
    \includegraphics[width=\linewidth]{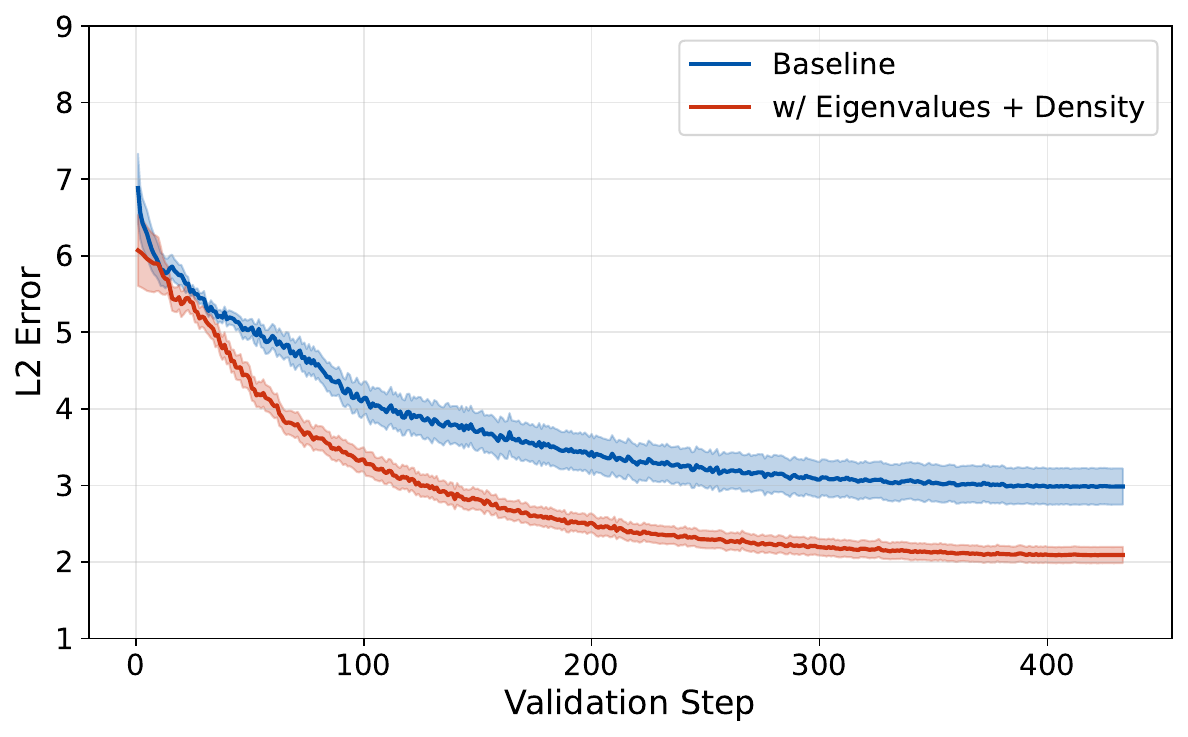}
    \caption{rnn\_input\_dim = 4}
    \label{fig:toy:tinyidnet in=4}
  \end{subfigure}
  \begin{subfigure}[t]{0.45\textwidth}
    \centering
    \includegraphics[width=\linewidth]{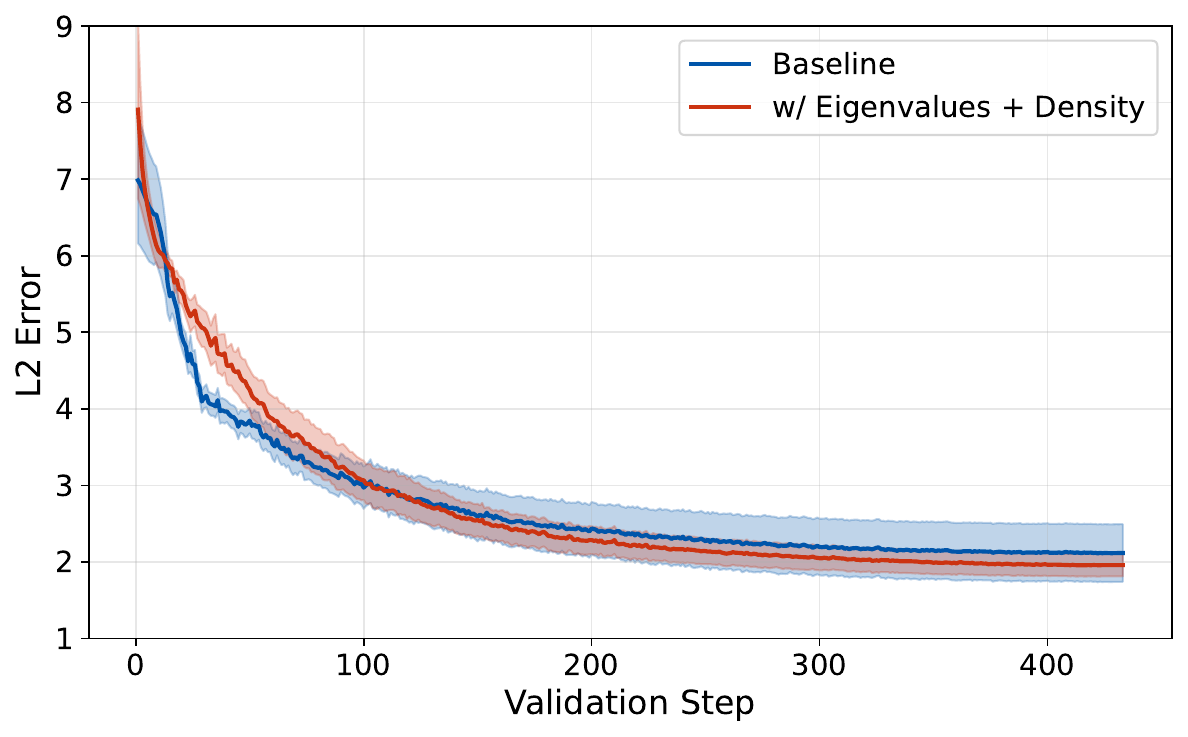}
    \caption{rnn\_input\_dim = 8}
    \label{fig:toy:tinyidnet in=8}
  \end{subfigure}
\caption{Test MSE error over training steps for \emph{Tiny IDNet}. We compare baseline features against combined extended features for different RNN input dimensions (4 and 8). Shaded regions denote $\pm 1$ standard error across runs. Models with lower input dimensionality \nergis{Nergis: should we say `lower capacity' instead? links better to the intro etc.} benefit more from the added eigenvalue and density features.}
  \label{fig:toy:tinyidnet}
\end{figure}

\noindent\textbf{Feature extending on Tiny IDNet}
We study feature extending under limited model capacity by varying Tiny IDNet's RNN input dimensionality. 
\Cref{fig:toy:tinyidnet} shows that extending the \originalcase{} inputs with \eigvalues{} and \filtervalues{}s consistently reduces MSE error, with larger gains at smaller input dimension. As capacity increases, the gap between baseline and extended inputs diminishes, suggesting that higher-capacity models can partially learn similar representations implicitly.
This suggests that the proposed features are particularly effective for low-complexity, resource-constrained models.


\noindent\textbf{Effect of texture on toy dataset with Tiny IDNet}
We repeat the texture-based analysis in \cref{sec:exp subsec:mlp-knn}, using Tiny IDNet.
We choose the textures from DTD~\cite{cimpoi14describing}, and train Tiny IDNet with  \texttt{rnn\_input\_dim}=4 and \texttt{hidden\_dim}=4. 
\Cref{tab:toy:tinyidnet:texture} reports the MSE error for different texture patterns, averaged over 5 independent runs.
Consistent with the earlier findings, extending the inputs with density and Harris eigenvalues improves flow estimation under textured conditions.
These results indicate that the proposed features generalize across other model architectures and are particularly beneficial when model capacity is limited.

\begin{table}[t]
\centering
\caption{Test MSE error for different texture patterns using \emph{Tiny IDNet} (\texttt{rnn\_input\_dim}=4, \texttt{hidden\_dim}=4) with \originalcase{} inputs and \bothshortvalues{}. For each texture, the mean and standard deviation over 5 independent runs are reported.}
\label{tab:toy:tinyidnet:texture}
\begin{adjustbox}{width=\textwidth}%
\begin{tabular}{lcccccc}
\hline
Feature Type & flecked-0103 & honeycombed-0139 & spiralled-0137 & stratified-0106 & woven-0053 & Overall (Mean $\pm$ Std) \\
\hline
Baseline & $5.5659 \smallpm{0.2523}$ & $4.8770 \smallpm{0.7717}$ & $4.4550 \smallpm{0.7552}$ & $5.0143 \smallpm{0.6433}$ & $4.7696 \smallpm{1.2014}$ & $4.9364 \smallpm{0.8663}$ \\
w/ Eig. + Density & $3.7602 \smallpm{0.9523}$ & $3.6062 \smallpm{0.4699}$ & $3.5789 \smallpm{1.0642}$ & $3.9544 \smallpm{0.5902}$ & $3.6059 \smallpm{0.8819}$ & $3.7094 \smallpm{0.8591}$ \\
Improvement(\%) & 32.44\% & 26.06\% & 19.67\% & 21.14\% & 24.40\% & 24.86\% \\
\hline
\end{tabular}
\end{adjustbox}
\end{table}

\begin{table}[t]
\caption{Optical flow results on the DSEC test set (compact version). IDNet 1/8 and 1/4 Res.\ refer to the model's internal resolution configuration. Sub-dataset results report mean$\pm$std over 3 runs. Best results for each category are shown in \textbf{bold}.}
\label{tab:dsec:paper_table_compact}
\begin{adjustbox}{width=\textwidth}%
\begin{tabular}{l l ccccc}
\hline
Category & Feature Type & $\mathrm{1PE}\downarrow$ & $\mathrm{2PE}\downarrow$ & $\mathrm{3PE}\downarrow$ & $\mathrm{EPE}\downarrow$ & $\mathrm{AE}\downarrow$ \\
\hline
\multirow{2}{*}{Sub-dataset, IDNet 1/8 Res.}
    & \Originalcase{}
        & $29.208\,\smallpm{0.518}$ & $10.805\,\smallpm{0.306}$ & $5.578\,\smallpm{0.132}$ & $1.191\,\smallpm{0.011}$ & $4.824\,\smallpm{0.046}$ \\
    & w/ Eig. + Density
        & $\mathbf{27.519}\,\smallpm{1.012}$ & $\mathbf{10.331}\,\smallpm{0.580}$ & $\mathbf{5.367}\,\smallpm{0.250}$ & $\mathbf{1.156}\,\smallpm{0.025}$ & $\mathbf{4.641}\,\smallpm{0.099}$ \\
\hline
\multirow{2}{*}{Full dataset, IDNet 1/8 Res.}
    & \Originalcase{}
        & 13.504 & 4.901 & 2.745 & 0.82  & 3.037 \\
    & w/ Eig. + Density 
        & \textbf{12.912} & \textbf{4.7}   & \textbf{2.645} & \textbf{0.805} & \textbf{2.968} \\
\hline
\multirow{2}{*}{Full dataset, IDNet 1/4 Res.}
    & \Originalcase{}
        & 10.256 & 3.448 & \textbf{1.928} & 0.725 & 2.755 \\
    & w/ Eig. + Density 
        & \textbf{9.868}  & \textbf{3.379} & 1.961 & \textbf{0.723} & \textbf{2.681} \\
\hline
\end{tabular}

\end{adjustbox}
\end{table}  

\subsection{DSEC dataset with IDNet}
\label{sec:exp subsec:dsec idnet}

We evaluate \eigvalues{} and \filtervalues{} on the standard real-world benchmark dataset DSEC~\cite{gehrig_dsec_2021} using the full IDNet architecture~\cite{wu_lightweight_2024}. 
Following~\cite{wu_lightweight_2024}, we use 4 deblurring iterations, event voxel representation with 15 bins, and Adam~\cite{kingma_adam_2017} with learning rate $1\times10^{-4}$ and a OneCycle learning rate schedule. We train the 1/8 resolution model for 150 epochs (batch size 5) and the 1/4 resolution model for 500 epochs (batch size 4).
We adopt the data augmentation protocol from~\cite{wu_lightweight_2024}, which includes random cropping to $288 \times 384$ pixels, vertical flips with probability 0.1, and horizontal flips with probability 0.5.

\Cref{tab:dsec:paper_table_compact} reports the results on the DSEC test set.
We compare the \originalcase{} features against extending with \eigvalues{} and \filtervalues{}s.
We evaluate using standard optical flow metrics: $n$-pixel error ($n$PE) denotes the percentage of pixels with endpoint error exceeding $n$ pixels, endpoint error (EPE) measures the average Euclidean distance between predicted and ground-truth flow vectors, and angular error (AE) quantifies the average angular deviation between predicted and ground-truth flow directions.

We first train on a subset of the DSEC dataset comprising approximately one-sixth of the full training data (see Supplementary for the list of sequences used).
For this sub-dataset experiment, we train at 1/8 resolution and report the average and standard deviation over 3 independent runs.
For the full dataset, we train the network at both 1/4 and 1/8 resolutions with and without the combined extended features.

The results demonstrate that extending with \eigvalues{} and \filtervalues{} consistently reduces errors across all experimental settings.
Notably, the improvement is higher on the sub-dataset, suggesting that the proposed features are beneficial in data-scarce scenarios.
Furthermore, the relative improvement at 1/8 resolution exceeds that at 1/4 resolution.
This aligns with our findings in~\cref{fig:toy:tinyidnet}, where we observed that lower-capacity models benefit more from the extended features.
We also observe that the angular error (AE) is reduced in both the sub-dataset and full dataset experiments.
This is consistent with our theoretical analysis in \cref{sec:method subsec:triangle theory}, where we established the connection between \eigvalues{} and motion direction through the structure tensor.
\nergis{I commented out the last part here, since I don't think we should use Fig. 1 as evidence to support any claim. The role of Fig. 1 is more to explain the idea in a simple way. The theory is the much better support.}
Additional evaluation on another real-world EVIMO2v2 dataset~\cite{burner_evimo2_2022}, 
an ablation of hyperparameters of extended features, and an analysis of their memory footprint and per-event computational cost are provided in the supplementary material.

\section{Conclusion}
\label{sec:conclusion}

We presented a study revealing that two features used for corner detection---the Harris eigenvalues of the structure tensor and the spatiotemporal density values---are in fact \emph{motion cues}, a property unique to event cameras with no analogue in frame-based vision.
We formally established the connection between Harris eigenvalues and motion direction for translating corners, showing that eigenvalues, combined with local geometry, can recover the direction of motion.
Through controlled experiments on synthetic data, we confirmed that extending local geometric features with eigenvalues and density values provides complementary motion information, that the improvements stem from informative features rather than increased model capacity, and that these features are robust to foreground textures and shot noise.
Integration into the state-of-the-art IDNet architecture on the real-world DSEC benchmark yielded consistent improvements in optical flow accuracy, with the largest gains observed in data-scarce training regimes and for lower-capacity models.
Our lightweight, plug-and-play approach requires no additional encoders or complex fusion strategies, making it easy to adopt in existing event-based pipelines.


\fi
\clearpage
\putbib[main,references]
\end{bibunit}
\makeatother

\ifsupplementary
\else
\clearpage
\begin{center}
{\Large\bfseries Supplementary Material}
\end{center}
\renewcommand{\theHsection}{appendix.\Alph{section}}
\renewcommand{\theHsubsection}{appendix.\Alph{section}.\arabic{subsection}}
\makeatletter
\begin{bibunit}[splncs04]

\clearpage
\putbib[main,references]
\end{bibunit}
\makeatother
\fi


\end{document}